\documentclass[12pt]{article}
\usepackage{natbib}

\input{preamble}

\usepackage{booktabs}

\begin{document}

\title{\bf Correlated Random Measures}

\author{Rajesh Ranganath\\
  Department of Computer Science\\Princeton University\\ \texttt{rajeshr@cs.princeton.edu}
 \\ \\
  David M. Blei\\
  Departments of Computer Science and Statistics\\ Columbia
  University \\  \texttt{david.blei@columbia.edu}
}
\date{}

\maketitle

\begin{abstract}

We develop correlated random measures, random measures where the atom
weights can exhibit a flexible pattern of dependence, and use them
to develop powerful hierarchical Bayesian nonparametric models.
Hierarchical Bayesian nonparametric models are usually built from
completely random measures, a Poisson-process based construction in
which the atom weights are independent.  Completely random measures
imply strong independence assumptions in the corresponding
hierarchical model, and these assumptions are often misplaced in
real-world settings.  Correlated random measures address this
limitation. They model correlation within the measure by using a
Gaussian process in concert with the Poisson process. With correlated
random measures, for example, we can develop a latent feature model
for which we can infer both the properties of the latent features and
their dependency pattern.  We develop several other examples as well.
We study a correlated random measure model of pairwise count data. We
derive an efficient variational inference algorithm and show improved
predictive performance on large data sets of documents, web clicks,
and electronic health records.

\end{abstract}

\section[Introduction]{Introduction}

Hierarchical Bayesian nonparametric models~\citep{Teh:2010a} have
emerged as a powerful approach to analyzing complex
data~\citep{Williamson:2010b, Fox:2011, Zhou:2012}.  These models
assume there are a set of patterns, or components, that underlie the
observed data; each data point exhibits each component with different
non-negative weight; the number of components is unknown and new data
can exhibit still unseen components.  Given observed data, the
posterior distribution reveals the components (including how many
there are), reveals how each data point exhibits them, and allows for this
representation to grow as more data are seen.  These kinds of
assumptions describe many of the most common hierarchical Bayesian
nonparametric models, such as the hierarchical Dirichlet
process~\citep{Teh:2006}, the Gamma-poisson
process~\citep{Titsias:2008}, the beta-Bernoulli
process~\citep{Thibaux:2007}, and others.

For example, in Section~\ref{sec:experiments} we analyze patient data from a
large hospital; each patient is described by the set of diagnostic
codes on her chart.  Potentially, the full data set reflects patterns
in diagnostic codes, each pattern a set of diagnoses that often occurs
together.  Further, some patients will exhibit multiple
patterns---they simultaneously suffer from different clusters of
symptoms.  With these data, a Bayesian nonparametric model can uncover
and characterize the underlying pattern and describe each patient in
terms of which patterns she exhibits. Recent innovations in
approximate posterior inference let us analyze such data at large
scale, uncovering useful characterizations of disease and injury for
both exploration and prediction. In our study on medical data, we
discover components that summarize conditions such as congestive
heart failure, diabetes, and depression (Table~\ref{tab:mayo-topics}).

But there is a limitation to the current state of the art in Bayesian
nonparametric models.  To continue with the example, each patient is
represented as an infinite vector of non-negative weights, one per
component.  (There is a countably infinite number of components.)  Most
hierarchical Bayesian nonparametric models assume that these weights
are uncorrelated---that is, the presence of one component is unrelated
to the presence (or absence) of the others. But this assumption is
usually unfounded.  For example, in the medical data we find that
type 2 diabetes is related to congestive heart failure (Table~\ref{tab:mayo-corr}).

In this paper we solve this problem. We develop correlated random
measures, a general-purpose construction for infusing covariance into
the distribution of weights of both random measures and
hierarchical Bayesian nonparametric
models. Our approach can capture that a large positive weight for one
component might covary with a large positive weight in another, a type
of pattern that is out of reach for most hierarchical Bayesian
nonparametric models.  We demonstrate that bringing such correlations
into the model both improves prediction and reveals richer exploratory
structures.  Correlated random measures can be used as a model for a
collection of observed weighted point processes and can be adapted to
a wide variety of proven Bayesian nonparametric settings, such as
language modeling~\citep{Teh:2006a}, time series
analysis~\citep{Fox:2011}, dictionary learning~\citep{Zhou:2009}, and
nested models~\citep{Paisley:2015}.

How do we achieve this? Most Bayesian nonparametric models are built
on completely random measures~\citep{Kingman:1967} and the
independence of the weights is an artifact of this
construction. To create correlated random measures, we infuse a
Gaussian process~\citep{Rasmussen:2005} into the construction with a
latent kernel between components. This lets us relax the strict
independence assumptions.  The details involve showing how to use the
Gaussian process in concert with the Poisson process, and without
sacrificing the technicalities needed to define a proper random
measure. As a result of the general construction, we can build
correlated variants of many hierarchical Bayesian nonparametric
models.

We will describe four correlated random measures.  The first is a
correlated nonparametric version of Poisson factorization
~\citep{Canny:2004, Gopalan:2014}.  This is a model of count data, organized in a
matrix, and it will be the model on which we focus our study.  We show
how to derive an efficient variational inference algorithm to
approximate the posterior and use it to analyze both medical data and
text data.  We also describe a correlated analog of the beta
process~\citep{Hjort:1990} and two correlated binary latent feature
models, each expanding on the hierarchical beta-Bernoulli
process~\citep{Griffiths:2006a,Thibaux:2007}.  We note that the
discrete infinite logistic normal model in~\citet{Paisley:2012b} is a
normalized correlated random measure, a correlated adaptation of
the hierarchical Dirichlet process~\citep{Teh:2006}.

\parhead{Related work.}  Correlated random measures (CorrRMs) can capture
general covariance between the measure of two sets, while also being
atomic and extendible to hierarchical models that share atoms.  In the
Bayesian nonparametric literature, researchers have proposed several
other random measures with covariance.  We survey this work.

\citet{Cox:1980} introduced the Cox process, a Poisson random measure
whose mean measure is also stochastic.  Cox processes can capture
covariance if the stochastic mean measure exhibits covariance.  Unlike
CorrRMs, however, Cox processes do not allow for noninteger atom
weights.  Furthermore, most common examples of Cox processes, such as
the log-Gaussian Cox process \citep{Moller:1998}, do not allow for
atom sharing.  We note that CorrRMs share the doubly stochastic
construction of the Cox process; in \myappendix{laplace-functional} we
show that CorrRMs can be alternatively viewed as a stochastic
transformation of a Poisson process.

Determinantal point processes \citep{Borodin:2009} are point processes
where the number of points is related to the determinant of a kernel
function.  Determinantal point processes exhibit ``anti-clumping''
(i.e., negative correlation) because atoms that are close together
will not appear together.  In contrast, hierarchical correlated random
measures do not rely on the atom values to determine their
correlation, and can capture both negative and positive correlation
among their weights.

\citet{Doshi-Velez:2009c} present a specific correlated feature model
by positing a higher level grouping of features.  In their model,
observations exhibit correlation through these groups.  However,
groups only contain features and thus these feature can only express
positive correlation.  In the latent feature models based on CorrRMs,
the latent locations of two features can induce a negative correlation
between their co-occurrence.

\citet{Ammann:1978} study infinitely divisible random measures 
that are not completely random. These measures are 
referred to as having ``aftereffects" in
that every atom in the random measure has more effect on the measure
than just its size. Correlated random measures are random measures with 
aftereffects, but they are not necessarily infinitely divisible.

As we have mentioned, the discrete infinite logistic normal (DILN)
\citep{Paisley:2012b}, a Bayesian nonparametric topic model, is a
normalized instance of a correlated random measure.  DILN first
generates top level shared atoms from a Dirichlet process, along with
latent locations for each.  It then draws each document with a gamma
process from those atoms and a Gaussian process evaluated at their
locations. Finally, it convolves these processes and normalizes to
form a probability measure.  We discuss DILN in detail in
Section~\ref{sec:DILN}.

Finally, there has been a lot of research in Bayesian nonparametrics
about dependent random measures, originating from the work of
\citet{MacEachern:1999}, broadly surveyed in \citet{Foti:2015}, and 
used in applications such as for dynamic ordinal data~\citep{Deyoreo:2015},
neuron spikes~\citep{Gasthaus:2009}, and images~\citep{Sudderth:2009}.  
Dependent random measures select atoms for
each observation through {\it a priori} covariates, such as a
timestamp associated with the observation.  Atoms are correlated, but
only though these observed covariates. The main ideas behind
correlated random measures and dependent random measures are
different.  Correlations in CorrRMs are not based on side information,
but rather are recovered through a random function associated with
each observation.  One dependent random measure that is close in
construction to correlated random measures is the dependent
Poisson process thinning measure of~\citet{Foti:2013}.  This measure
can be reinterpreted as a type of correlated random measure; we
discuss this connection with technical details in Section~\ref{sec:crm}. Another
construction, compound 
random measures~\citep{Griffin:2014},
builds dependent random measures 
by using a score functions to generate a set of measures conditional
on a shared Poisson process. Compound random measures and the dependent
Poisson process thinning measure share with our approach the idea of 
separating out the atom generation from the independence breaking portion.

\section{Background: Completely Random Measures}
\label{sec:background}

In this section we review random
measures~\citep{Kingman:1967,Cinlar:2011}.  We describe the Poisson
random measure, completely random measures, and normalized random
measures.  This sets the stage for our construction of the correlated
random measure in Section~\ref{sec:crm}.

A random measure $M$ is a stochastic process that is indexed by a
sigma algebra.  Let $(E, \cE)$ be a measurable space, for example $E$
is the real line and $\cE$ are the Borel sets.  A random measure is a
collection of random variables $M(A) \in [0, \infty]$, one for each
set $A \in \cE$.  The expectation of a random measure is called the
mean measure, which we denote $\nu(A) \triangleq \E[M(A)]$.

One subclass of random measures is the class of completely random
measures~\citep{Kingman:1967}.  A completely random measure is a
random measure $M(\cdot)$ such that for any disjoint finite collection
of sets $A_1, A_2, \ldots, A_n$, the corresponding realizations of the
measure on those sets $M(A_1), M(A_2), \ldots, M(A_n)$ are independent
random variables.  Completely random measures encompass many of the
constructions in Bayesian nonparametric statistics.  Some examples
include the Poisson process, the beta process~\citep{Hjort:1990}, the
Bernoulli process~\citep{Thibaux:2007}, and the gamma
process~\citep{Ferguson:1973}.

We begin by describing the simplest example of a completely random
measure, the Poisson random measure.  The Poisson random measure is
constructed from a Poisson process.  It is characterized solely by its
mean measure $\nu(\cdot): \cE \rightarrow [0, \infty]$, which is an
arbitrary measure on $(E, \cE)$.  The complete characterization of a
Poisson random measure $M(\cdot)$ is that the marginal distribution of
$M(A)$ is a Poisson with rate $\nu(A)$.

We represent a Poisson random measure with a set of atoms $a_i$ in $E$
and a sum of delta measures on those atoms~\citep{Cinlar:2011},
\begin{equation*}
  M(A) = \sum_{i=1}^{\infty} \delta_{a_i}(A).
\end{equation*}
The delta measure $\delta_{a_i}(A)$ equals one when $a_i \in A$ and
zero otherwise.  Note there can be a countably infinite set of atoms,
but only if $\nu(E) = \infty$.\footnote{This fact follows from the
  marginal distribution $M(E) \sim \textrm{Poisson}(\nu(E))$ and that
  a Poisson random variable with rate equal to $\infty$ is $\infty$
  almost surely.}  The distribution of the atoms comes from the mean
measure $\nu(\cdot)$.  For each finite measurable set $A$, the atoms
in $A$ are distributed according to $\nu(\cdot) / \nu(A)$.

We now expand the simple Poisson random measures to construct more
general completely random measures.  Consider a Poisson process on the
cross product of $E$ and the positive reals, $E \times \mathbb{R}_+$.
It is represented by a set $\{(a_i, w_i)\}_{i=1}^{\infty}$; each pair
contains an atom $a_i$ and corresponding weight
$w_i \in \mathbb{R}_+$. The completely random measure is
\begin{equation}
  \label{eq:completely_random_measure}
  M(A) = \sum_{i=1}^{\infty} w_i \delta_{a_i}(A).
\end{equation}
This Poisson process is characterized by its mean measure, called the
Levy measure, which is defined on the corresponding cross product of
sigma algebras,
$$\nu(\cdot, \cdot): \mathcal{E} \times \mathcal{B}(\mathbb{R_+}) \rightarrow [0, \infty].$$
We note that completely random measures also have fixed components,
where the atoms are fixed in advance and the weights are random.  But
we will not consider fixed components here.

We call the process homogenous when the Levy measure factorizes,
$\nu(A, R) = H(A) \hat{\nu}(R)$; we call $H$ the base measure.  For
example, in a nonparametric mixture of Gaussians the base measure is a
distribution on the mixture locations~\citep{Escobar:1995}; in a
nonparametric model of text, the base distribution is a Dirichlet over
distributions of words~\citep{Teh:2006}.

We confirm that $M(\cdot)$ in \myeq{completely_random_measure} is a
measure.  First, $M(\emptyset) = 0$.  Second, $M(A) \geq 0$ for any $A$.
Finally, $M(\cdot)$ satisfies countable additivity.  Define $A$ to be
the union of disjoint sets $\{A_1, A_2, \ldots \}$.  Then
$M(A) = \sum_k M(A_k)$.  This follows from a simple argument,
\begin{equation*}
  \label{eq:countable-additivity}
  \sum_{k=1}^{\infty} M(A_k) = \sum_{k=1}^{\infty} \sum_{i=1}^{\infty}
  w_i \delta_{a_i}(A_k) = \sum_{i=1}^{\infty} w_i \sum_{k=1}^{\infty}
  \delta_{a_i}(A_k) = \sum_{i=1}^{\infty} w_i \delta_{a_i}(A) = M(A).
\end{equation*}
We used Tonelli's theorem to interchange the summations.

One example of a completely random measure is the gamma
process~\citep{Ferguson:1973}.  It has Levy measure
\begin{equation*}
  \nu(da, dw) \triangleq H(da)e^{-cw}/wdw.
\end{equation*}
This is called the gamma process because if $M \sim \gammapro(H, c)$
the random measure $M(A)$ on any set $A \in \cE$ is gamma distributed
$M(A) \sim \textrm{Gamma}(H(A), c)$, where $H(A)$ is the shape and $c$
is the rate~\citep{Cinlar:2011}.  The gamma process has an infinite
number of atoms---its Levy measure integrates to infinity---but the
weights of the atoms are summable when the base measure is finite
($H(E) < \infty$) because $\E[M(E)] = \frac{H(E)}{c}$.  Finally, when
$M(E) < \infty$, we can normalize a completely random measure to
obtain a random probability measure. For example, we construct the
Dirichlet process~\citep{Ferguson:1973} by normalizing the gamma
process.

\section{Correlated Random Measures}
\label{sec:crm}

The main limitation of a completely random measure is articulated in
its definition---the random variables $M(A_i)$ are independent.
(Because they are normalized, random probability measures exhibit some
negative correlation between the $M(A_i)$, but cannot capture other
types of relationships between the probabilities.)  This limitation
comes to the fore particularly when we see repeated draws of a random
measure, such as in hierarchical Bayesian nonparametric
models~\citep{Teh:2010a}.  In these settings, we may want to capture
and infer a correlation structure among $M(A_i)$ but cannot do so with
the existing methods (e.g., the hierarchical Dirichlet process).  To
this end, we construct correlated random measures.  Correlated random
measures build on completely random measures to capture rich
correlation structure between the measure at disjoint sets, and this
structure can be estimated from data.

We built completely random measures from a Poisson process by
extending the space from simple atoms (in the Poisson process) to the
space of atoms and weights (in a completely random measure).  We build
correlated random measures from completely random measures by
extending the space again.  As for a completely random measure, there
is a set of atoms and uncorrelated weights.  We now further supply
each tuple with a ``location'', a vector in $\bbR^d$, and extend the
mean measure of the Poisson process appropriately. A correlated random
measure is built from a Poisson process on the extended space of
tuples $\{(a_i, w_i, \ell_i)\}_{i=1}^{\infty}$.

In the completely random measure of Equation~\ref{eq:completely_random_measure},
the uncorrelated weights $w_i$ give the measure at each atom.  In a
correlated random measure there is an additional layer of variables
$x_i$, called the \textit{transformed weights}.  These transformed
weights depend on both the uncorrelated weights $w_i$ and a random
function on the locations $F(\ell_i)$.  In the random measure, they
are used in place of the uncorrelated weights,
\begin{align}
  \label{eq:corr-rm}
  M(A) = \sum_{i=1}^{\infty} x_i \delta_{a_i}(A).
\end{align}
It is through the random function $F(\cdot)$, which is drawn from a
Gaussian process~\citep{Rasmussen:2005}, that the weights exhibit
correlation.

We first review the Gaussian process (GP) and then describe how to
construct the transformed weights.  A Gaussian process is a random
function $F(\ell_i)$ from $\bbR^d \rightarrow \bbR$.  It is specified
by a positive-definite kernel function\footnote{This means that for a
  finite collection of inputs, the kernel produces a positive definite
  matrix.}  $K(\ell_i, \ell_j)$ and mean function $\mu(\ell_i)$.  The
defining characteristic of a GP is that each joint distribution of a
collection of values is distributed as a multivariate normal,
\begin{equation*}
  (F(\ell_1), \dots, F(\ell_n)) \sim \cN(m, \Sigma),
\end{equation*}
where $m_i = \mu(\ell_i)$ and $\Sigma_{ij} = K(\ell_i, \ell_j)$.

In a correlated random measure, we draw a random function from a GP,
evaluate it at the locations of the tuples $\ell_i$, and use these
values to define the transformed weights.  We specify the
\textit{transformation distribution} of $x_i \in \bbR^+$ denoted
$T(x_i \g w_i, F(\ell_i))$.  It depends on both the uncorrelated
weights $w_i$ and the GP evaluated at $\ell_i$. For example, one
transformation distribution we will consider below is the gamma,
\begin{equation}
  \label{eq:transformation-gamma}
  x_i \sim \textrm{Gamma}(w_i, \exp\{-F(\ell_i)\}).
\end{equation}
But we will consider other transformation distributions as well.  What
is important is that the $x_i$ are positive random variables, one for
each atom, that are correlated through their dependence on the GP $F$.

We have now fully defined the distribution of the transformed weights
$x_i$ that are used in the correlated random measure of
Equation~\ref{eq:corr-rm}.  We emphasize that in a completely random measure
the weights are independent. The arguments that $M(\cdot)$ is a measure,
however, only relied on its form, and not on the independence of the
weights. (See Equation~\ref{eq:countable-additivity}.)

In summary, we build a correlated random measure by specifying the
following: the mean measure of a Poisson process on atoms, weights,
and locations $\nu(da, dw, d\ell)$; a kernel function
$K(\ell_i, \ell_j)$ between latent locations and a mean function
$m(\ell_i)$; and the conditional transformation distribution
$T(\cdot \g w_i, F(\ell_i))$ over positive values.  With these
elements, we draw a correlated random measure as follows:
\begin{eqnarray}
  \{(a_i, w_i, \ell_i)\}_{i=1}^{\infty} &\sim& \poissonpro(\nu) \label{eq:step-one}\\
  F &\sim& \gaussianpro(m, K) \label{eq:step-two} \\
  x_i &\sim& T(\cdot \g w_i, F(\ell_i)) \label{eq:step-three}.
\end{eqnarray}
The random measure $M(\cdot)$ is in Equation~\ref{eq:corr-rm}.

Before turning to some concrete examples, we set up some useful
notation for correlated random measures.  We denote the infinite set
of tuples from the Poisson process (Equation \ref{eq:step-one}) with
\begin{equation*}
  \mathcal{C} \triangleq \{(a_i, w_i, \ell_i)\}_{i=1}^{\infty}.
\end{equation*}
Given these tuples, the process for generating a correlated random
measure first draws from a Gaussian process (Equation \ref{eq:step-two}), then
transforms the weights (Equation \ref{eq:step-three}),
and finally constructs the measure from an infinite sum
(Equation~\ref{eq:corr-rm}).  We shorthand this process with
$M \sim \corr(\mathcal{C}, K, \mu, T)$.\footnote{As in the
  construction of completely random measures, correlated random
  measure can also include fixed components, where the tuples
  are fixed, but the $x_i$ are random.}

We note that correlated random measures generalize completely random
measures.  Specifically, we can construct a completely random measure
from a correlated random measure by setting the mean measure
$\nu(dx, dw, d\ell)$ to match the corresponding completely random measure
mean measure
$\nu(dx, dw)$ (i.e., the location distribution does not matter) and
asserting that $x_i = w_i$ with probability one.

With the full power of the correlated random measure, we can construct
correlated versions of common random measures such as the gamma
process, the beta process, and normalized measures such as the
Dirichlet process.  We give two examples below.

\parhead{Example: Correlated Gamma process.} We discussed the gamma
process as an example of a completely random measure. We now extend
the gamma process to a correlated gamma process.  First, we must
extend the mean measure to produce atoms, weights, and locations.  We
specify an additional distribution of locations $L(\ell)$---we
typically use a multivariate Gaussian---and expand the mean measure of
the gamma process to a product,
\begin{equation*}
  \nu(da, dw, d\ell) = L(d\ell) H(da) e^{-cw}/w dw.
\end{equation*}
Second, for the transformation distribution, we choose the gamma in
Equation~\ref{eq:transformation-gamma}.  Finally, we define the GP parameters,
the kernel $K(\ell_i, \ell_j) = \ell_i^\top\ell_j$ and a zero mean
$\mu(\ell_i) = 0$.  With these components in place, we draw from
Equation \ref{eq:step-one}, Equation \ref{eq:step-two}, and Equation \ref{eq:step-three}.  This is one
example of a correlated random measure.

\parhead{Example: Correlated Dirichlet process.}  We can normalize the
measure to construct a \textit{correlated random probability measure}
from a correlated random measure.  If $M(\cdot)$ is a correlated
random measure, then
\begin{equation*}
  G(A) = M(A) / M(E)
\end{equation*}
is a correlated random probability measure.  As we discussed in
\mysec{background}, the Dirichlet process is a normalized gamma
process~\citep{Ferguson:1973}.  When we normalize the correlated gamma
process, we obtain a correlated Dirichlet process.

This construction requires that $M(E) < \infty$. Proposition 2 in
\myappendix{integrability} describes conditions for well-defined
normalization (i.e., $M(E) < \infty$) in correlated random
measures. Roughly, these conditions require finiteness of the expected
value of the transformed weights against the product of the
transformation distribution and the Levy measure.  This condition
plays a central role in constructing useful Bayesian nonparametric
models.

\parhead{The correlation structure.}  Finally, we calculate the
correlation structure of a correlated random measure.  Consider a
measure,
$M \sim \corr(\mathcal{C}, K, m, T)$.  To understand the nature of the
correlation, we compute $\textrm{Cov}(M(A), M(B))$ for two sets $A$
and $B$.

We express this covariance in terms of the covariance between atom
weights,
\begin{equation*}
  \textrm{Cov}(M(A), M(B)) =
  \sum_{i=1}^\infty \sum_{j=1}^\infty
  \textrm{Cov}(X_i, X_j) \delta_{a_i}(A) \delta_{a_j}(B).
\end{equation*}
In other words, the covariance between the measure of two sets is the
sum of the covariances between the transformed weights of the atoms in
the two sets.  In a completely random measure, the atom covariance is
zero except when $a_i = a_j$.  Thus its covariance depends on the
overlap between sets, and nothing else.  In a correlated random
measure, however, there may be non-zero covariance between the
transformed weights.

For now we are holding the underlying Poisson process $\mathcal{C}$
fixed, i.e., the atoms, untransformed weights, and locations.  The
covariance between transformed weights is
\begin{equation}
  \textrm{Cov}(X_i, X_j \g \mathcal{C}) = \E[X_i X_j] - \E[X_i]
  \E[X_j].
  \label{eq:conditional_covariance_1}
\end{equation}
These expectations are driven by two sources of randomness.  First
there is a Gaussian process $F$, a random function evaluated at the
fixed locations.  Second there is the transformation distribution $T$.
This is a distribution of an atom's transformed weight, conditional on
its untransformed weight and the value of the Gaussian process at its
location (see Equation~\ref{eq:step-three}).

Using iterated expectation, we write the conditional covariance in
Equation~\ref{eq:conditional_covariance_1} in terms of the conditional mean of the
transformed weights, $\mu_i \triangleq \E[X_i \g F(\ell_i), w_i]$.
This is a function of the Gaussian process $F$.  We rewrite the
conditional covariance,
\begin{equation}
  \textrm{Cov}(X_i, X_j \g \mathcal{C}) = \E[\mu_i \mu_j] - \E[\mu_i]
  \E[\mu_j],
  \label{eq:conditional_covariance}
\end{equation}
where the expectations are taken with respect to the Gaussian process.
For the first term, the distribution is governed by the distribution
of the pair $(F(\ell_i), F(\ell_j))$, which is a bivariate normal.
For the second term, the marginals are governed by $F(\ell_i)$ and
$F(\ell_j)$, which are univariate normal distributions.

\section{Hierarchical Correlated Random Measures}

A correlated random measure takes us from a set of tuples to a random
measure by way of a Gaussian process and a transformation
distribution.  When used in a downstream model of data, we can infer
the latent correlation structure from repeated realizations of
measures from the same set of tuples.  It is thus natural to build
\textit{hierarchical correlated random measures}.  Hierarchical
correlated random measures are the central use of this new
construction.

In a hierarchical correlated random measure, we first produce a set of
tuples $\{(a_i, w_i, \ell_i)\}_{i=1}^{\infty}$ from a Poisson process
and then re-use that set in multiple realizations of a correlated
random measure.  In each realization, we fix the tuples (weights,
atoms, and locations) but draw from the Gaussian process anew; thus we
redraw the transformed weights for each realization.

As for the simple correlated random measure, we first specify the mean
measure of the Poisson process $\nu(\cdot)$, the kernel and mean for
the Gaussian process $K(\cdot, \cdot)$, and the conditional
transformation distribution $T(\cdot \g w_i, G(\ell_i))$.  We then
draw $n$ hierarchical correlated random measures as follows:
\begin{eqnarray*}
  \mathcal{C} &\sim& \poissonpro(\nu) \\
  M_j(A) &\sim& \corr(\mathcal{C}, m, K, T).
\end{eqnarray*}

This is a hierarchical Bayesian nonparametric model~\citep{Teh:2010a}.
There are multiple random measures $M_j$.  Each shares the same set of
atoms, locations, and weights, but each is distinguished by its own
set of transformed weights.\footnote{In our empirical study of
  \mysec{experiments}, we will also endow each with its own mean
  function to the Gaussian process, $m_j(\cdot)$.  Here we omit this
  detail to keep the notation clean.}  The correlation structure of
these transformed weights is shared across measures.  We note that
this construction generalizes the discrete infinite logistic
normal~\citep{Paisley:2012b}, which is an instance of a normalized
correlated random measure.

We use this construction in a model of groups of observations $y_j$,
for which we must construct a likelihood conditional on the correlated
RM.  To construct a likelihood, many hierarchical Bayesian
nonparametric models in the research literature use the integral with
respect to the random measure.  (This is akin to an unnormalized
``expectation.'').  Define $Ma \triangleq \int a M(da)$, and note that
in a discrete random measure this integral is an infinite sum,
\begin{equation}
  \label{eq:dot-product}
  Ma = \sum_i x_i a_i.
\end{equation}
The $j$th observations are drawn from a distribution parameterized
from this sum, $y_j \sim p( \cdot \g Ma)$.  For example, we will study
models where $Ma$ is a collection of rates for independent Poisson
distributions.

We present several examples of hierarchical correlated random
measures. First, we develop \textit{correlated nonparametric Poisson
  factorization} (CNPF) for factorizing matrices of discrete data.
This is the example we focus on for posterior inference
(\mysec{inference}) and our empirical study (\mysec{experiments}).  We
then illustrate the breadth of correlated random measures with two
other examples, both of which are latent feature models that build
correlations into the class of models introduced by
\citet{Griffiths:2006a}.  Finally, we discuss the discrete infinite
logistic normal (DILN) of~\citet{Paisley:2012b}.  We show that DILN is
a type of normalized correlated random measure.

\subsection{Correlated Nonparametric Poisson Factorization}
\label{sec:CNPF}

Bayesian nonparametric Poisson matrix
factorization~\citep{Gopalan:2014} combines gamma
processes~\citep{Ferguson:1973} with Poisson likelihoods to factorize
discrete data organized in a matrix.  The number of factors is
unknown and is inferred as a consequence of the Bayesian
nonparametric nature of the model.

For concreteness we will use the language of \textit{patients} getting
\textit{diagnoses} (e.g., patients going to the hospital and getting marked
for medical conditions).
In these data, each cell of the matrix
$y_{uj}$ is the number of times patient $u$ was marked for diagnosis $j$.  The
goal is to factorize users into their latent ``health statuses'' and
factorize items into their latent ``condition groups''.  These inferences
then let us form predictions about which unseen codes a patient might
have. Though we focus our attention here on patients getting
diagnoses, we emphasize that discrete matrices are widely found in
modern data analysis problems.  In our empirical study, we will also
examine matrices of documents (rows) organized into word
counts (columns) from a fixed vocabulary and user (rows) clicks over
a fixed collection of items (columns).

We will use a hierarchical correlated random measure to model these
data, where each group is a patient and the group-specific data are her
vector of per-diagnosis counts.  An atom $a_i$ is a vector of positive
weights for each diagnosis, drawn from independent gamma distributions,
$H(a) = \textrm{Gamma}(\alpha, \beta)$.  When the posterior of these atoms is
estimated from diagnosis counts, they will represent semantic
groups of conditions such as ``diabetes," ``heart disease," or `cancer.''
Table \ref{tab:mayo-topics} displays some of the atoms inferred from
a set of patients from the Mayo Clinic.

In using a correlated random measure, the idea is that patients'
expression for these conditions are represented by the per-group weights
$x_i$.  Intuitively, these exhibit correlation.  A patient who has
``heart'' conditions is more likely to also have ``vascular'' conditions
than ``cancer.''  (To be clear, these groupings are the latent
components of the model.  There are an unbounded number of them, they
are discovered in posterior inference, and their labels are not
known.)  Using these correlations, and based on her history, a
correlated model should better predict which diagnoses a patient will have.

We now set up the model.  We set the mean measure for the shared atoms
to be
\begin{align*}
  \nu(da, dw, d\ell) \triangleq \textrm{Gamma}(da, \alpha, \beta)
  \textrm{Normal}(d\ell, 0, I_d \sigma_l^2)  e^{-cw}/w dw.
\end{align*}
We define the GP mean function to be a per-patient constant
$m_u(\ell_i) = \mu_u$, where $\mu_u \sim \mathcal{N}(0, \sigma_m^2)$.
These per-patient GP means account for data where some patients tend
to be sicker than others.  We define the GP kernel function to be
$K(\ell_i, \ell_j) = \ell_i^\top \ell_j$.

Finally, we consider two different transformation distributions. The
first transformation distributions is as in
Equation~\ref{eq:transformation-gamma},
\begin{align*}
  x_i \sim \textrm{Gamma}(w_i, \exp\{-F(\ell_i)\}).
\end{align*}
The second is
\begin{align*}
  x_i \sim \textrm{Gamma} \left(w_i, \frac{1}{\log(1 + \exp\{F(\ell_i)\}} \right),
\end{align*}
where $\log(1 + \exp(\cdot))$ is known as the softplus function.  With
these definitions, we can compute the conditional covariance for $x_i$
and $x_j$ using Equation~\ref{eq:conditional_covariance}. Table
\ref{tab:mayo-corr} displays some of positive correlations between
atoms found on patient diagnosis counts. These correlations are
captured by the locations, which are shared across patients, associated
with each atom. Atoms with positive covariance in this model will have
inferred locations that have a large inner product.

With these components in place, correlated nonparametric Poisson
factorization is
\begin{align*}
  \mathcal{C} &\sim \poissonpro(\nu) \\
  \mu_u &\sim \mathcal{N}(0, \sigma_m^2) \\
  M_u &\sim \corr(\mathcal{C}, \mu_u, K, T) \\
  y_u &\sim p(\cdot \g M_u a).
\end{align*}
The distribution of $y_u$ is a collection of Poisson variables, one for each
diagnosis $j$, where
\begin{equation}
  \label{eq:poisson-likelihood}
  y_{uj} \sim \textrm{Poisson}\left( \sum_{i=1}^\infty x_{ui} a_{ij} \right).
\end{equation}
Recall that the atoms $a_i$ are each a vector of gamma variables, one
per diagnosis, and so $a_{ij}$ is the value of atom $i$ for diagnosis
$j$. For this model to be well defined each rate in
Equation~\ref{eq:poisson-likelihood} must be finite. Using proposition 2 in
\myappendix{integrability}, it is finite almost surely if
$\sigma_l < 1$.

The sum that defines the rate of $y_{ui}$ is an infinite sum of patient
weights and condition weights.  Thus, this model amounts to a
factorization distribution for $y_{ui}$.  Given observed data, the
posterior distribution of the atoms $a_i$ and transformed patient
weights $x_i$ gives a mechanism to form predictions.  Note that the
atoms $a_i$ are shared across patients, but through $x_i$ each patient
exhibits them to different degree.  We discuss how to approximate this
posterior distribution in \mysec{inference}.

\subsection{Correlated latent feature models}

Mixture models, such at the Dirichlet process
mixture~\citep{Antoniak:1974}, are the most commonly used Bayesian
nonparametric model. In mixture models, each observation exhibits only
a single class. Many data, such as images of multiple objects, are
better characterized as belonging to multiple classes.  Latent feature
models posit that each observation is associated with some number of
latent classes, taken from a set of features shared by all
observations.  For each observation, its likelihood depends on
parameters attached to its active features (e.g., a sum of those
parameters).  Examples of latent feature models include factorial
mixtures \citep{Ghahramani:1995} and factorial hidden Markov models
\citep{Ghahramani:1997}.  (Latent feature models are closely connected
to spike and slab priors \citep{Ishwaran:2005}.)

Bayesian nonparametric latent feature models allow the number of
features to be unbounded. As an example, consider analyzing a large
data set of images.  Latent features could correspond to image
patches, such as recurring objects that appear in the images.  In
advance, we might not know how many objects will appear in the data
set.  BNP latent feature models attempt to solve this problem. BNP
latent feature models have been used in many domains such
as image denoising \citep{Zhou:2011} and link prediction in graphs \citep{Miller:2009}.

The most popular BNP latent feature model is the hierarchical
beta-Bernoulli process \citep{Thibaux:2007}.  This process was
originally developed as the Indian Buffet process, which marginalized
out the beta process~\citep{Griffiths:2006a}.  Before developing the
correlated version, we review the beta-Bernoulli process.

The beta process is a completely random measure with atom weights in
the unit interval (0,1).  Its Levy measure is
\begin{align*}
  \nu(da, dw) =  H(da) \alpha w^{-1} (1 - w)^{\alpha - 1}.
\end{align*}
We use the beta process in concert with the Bernoulli process, which
is a completely random measure parameterized by a random measure with
weights in the unit interval, i.e., a collection of atoms and corresponding weights. A draw
from a Bernoulli process selects each atom with probability equal to
its weight.  This forms a random measure on the underlying space,
where each weight is one or zero (i.e., where only a subset of the
atoms are activated).  Returning to latent feature models, the
beta-Bernoulli process is
\begin{eqnarray*}
  B &\sim& \textrm{Beta-Process}(H, \alpha) \\
  B_n &\sim& \textrm{Bernoulli-Process}(B) \\
  y_n &\sim& p(\cdot \g B_n)
\end{eqnarray*}
The beta process generates the feature atoms; the Bernoulli processes
chooses which features are active in each observation.

This model is built on completely random measures.  Thus, the
appearances of features in each observation are independent of one
other.  Correlated random measures relax this assumption.  Consider a
latent feature model of household images with image patch
features. The completely random assumption here implies that the
appearance of a spoon is independent of the appearance of a fork.  Our
construction can account for such dependencies between the latent
features.  Below we will give two examples of correlated nonparametric
latent feature models, one based on the beta process and the other
based on the gamma process.

One method to develop a correlated beta-Bernoulli process is to
define transformed weights at the Bernoulli process level. We define
the transformation distribution to be
\begin{eqnarray*}
x_{ni} \sim \textrm{Bernoulli}(\sigma(\sigma^{-1}(w_i) + F(\ell_i))),
\end{eqnarray*}
where $\sigma$ is the sigmoid function $\sigma(x) = 1/(1 + \exp\{-x\})$.  Thus
the beta-Bernoulli correlated latent feature model is
\begin{eqnarray*}
  \mathcal{C} &\sim& \poissonpro(H(da) L(d\ell) \alpha w^{-1} (1 - w)^{\alpha - 1}) \\
  M_n &\sim& \corr(\mathcal{C}, \mu, K, T).
\end{eqnarray*}
(We defer defining $\mu$ and $K$, as they will be application specific.)

We do not need to use the beta process to define a correlated latent
feature model; what is important is that the per-observation weights
are either one or zero.  For example, if the top level process is a
gamma process, which produces positive weights, then we can define the
transformation distribution to be
\begin{eqnarray*}
  x_{ni} \sim \textrm{Bernoulli}\left(\frac{w_i \exp(F(\ell_i))}{1 +
  w_i \exp(f(\ell_i))}\right).
\end{eqnarray*}
The resulting gamma-Bernoulli correlated latent feature model is
\begin{eqnarray*}
  \mathcal{C} &\sim& \poissonpro(H(da) L(d\ell)  e^{-cw}/w dw) \\
  M_n &\sim& \corr(\mathcal{C}, \mu, K, T).
\end{eqnarray*}

The beta-Bernoulli process uses only a finite number of features
to generate a finite number of observations.
In \myappendix{finiteness}, we give some
conditions under which the correlated latent feature models do the
same.

\subsection{Discrete infinite logistic normal}
\label{sec:DILN}

The correlated random measure construction that we developed
generalizes the discrete infinite logistic normal (DILN)
\citep{Paisley:2012b}.  DILN is an example of a normalized hierarchical
correlated random measure; its atom weights come from a normalized
gamma random measure, i.e., a Dirichlet process.

DILN was developed as a Bayesian nonparametric mixed-membership model,
or topic model, of documents.  In DILN, each document mixes a set of
latent topics (distributions over terms), where the per-document topic
proportions can exhibit arbitrary correlation structure.  This is in
contrast to a hierarchical Dirichlet process topic
model~\citep{Teh:2006}, where the topic proportions are nearly
independent.

We will express DILN in terms of a correlated random measure.  The
observations $w_{uj}$ are categorical variables, i.e., word $j$ in
document $u$.  Set the kernel
$K(\ell_i, \ell_j) = \ell_i^\top \ell_j$, and set $\alpha$ and $\beta$
to be positive hyperparameters.  Set the transformation distribution
to be
\begin{align*}
x_i \sim \textrm{Gamma}(\beta w_i, \exp\{-F(\ell_i)\}).
\end{align*}
With our construction, DILN is
\begin{eqnarray*}
  \mathcal{C} &\sim& \textrm{Dirichlet-Process}(\alpha, H(da) \times
                     \mathcal{N}(d\ell, 0, \sigma_l^2 I_d)) \\
  M_u &\sim& \textrm{Normalized-}\corr(\mathcal{C}, 0, K, T) \\
  z_{uj} &\sim& M_u \\
  w_{uj} &\sim& z_{uj}.
\end{eqnarray*}
Note that the shared tuples come from a Dirichlet process, i.e., a
normalized gamma process.  When modeling documents, the base
distribution over atoms $H(da)$ is a Dirichlet distribution over the
vocabulary.

This is a mixed-membership model---there is an additional layer of
hidden variables $z_{uj}$, drawn from the random measure, before
drawing observations $w_{uj}$.  These hidden variables $z_{ui}$ will
be atoms, i.e., distributions over the vocabulary, from the set of
shared tuples and drawn with probability according to the per-document
transformed weights.  Each observation $w_{uj}$ is drawn from the
distribution over terms given in its atom $z_{uj}$.

\citet{Paisley:2012b} show the normalization step is well defined when
$\sigma_l < 1$.  Viewing DILN through the lens of correlated random
measures makes clear what can be changed.  For example, the top level
choice of the Dirichlet process is not critical. It could be any
random measure that places finite total mass, such as a gamma process
or a beta process.

\subsection{Connection to Dependent Random Measures}
Finally, we discuss the detailed connection between correlated random
measures and dependent random measures~\cite{MacEachern:1999}.
Dependent random measures are a collection of measures indexed
by covariates. A broad class of dependent random 
measures can be created by thinning a Poisson process \citep{Foti:2013}.
Given a draw from a Poisson process
$(a_i, w_i, \ell_i)_{i=1}^\infty$, where $a$ are atoms, $\ell$ are locations
in the covariate space, and $w$ are weights, the thinned dependent
random measure $B$ for user $u$ with covariate $\theta_u$ is
 \begin{eqnarray}
 B_u(A) &=& \sum_{i=1}^\infty  x_{ui} \delta_{a_i} (A)
\nonumber \\
 x_{ui} &\sim& w_i\textrm{Bernoulli}(k(\theta_u, \ell_i)), \nonumber
 \end{eqnarray}
 where $k$ is a function from $T \times L \to [0, 1]$. This construction is related to CorrRMs.
 Consider the correlated random measure
 \begin{eqnarray}
 x_{ui} &\sim& w_i\textrm{Bernoulli}(\sigma(F(\ell_i))) \nonumber
 \\
 M_u(A) &\sim& \corr((a_i, w_i, \ell_i)_{i=1}^\infty, m, K, T). \nonumber
 \end{eqnarray}
 From, this we can see that
 $B_u(A) \edist M_u(A)$ when $\sigma(F_u(\ell_i)) = k(\theta_u, \ell_i)$.  In other words,
 thinned dependent random measures are equivalent to a type of correlated
 random measure where the random function, $F_u$ associated with each user
 is known and given by the covariate. 

We note that dependent random measures map from covariates to
measures.  Thus they can be viewed as a type of measure-valued
regression. In parallel, correlated random measures use latent
covariates.  In this sense, they can be viewed as measure-valued

\section{Variational Inference for Correlated Nonparametric Poisson Factorization}
\label{sec:inference}

Computing the posterior is the central computational problem in
Bayesian nonparametric modeling. However, computing the posterior
exactly is intractable. To approximate it, we use variational
inference \citep{Jordan:1999,Wainwright:2008}, an alternative to
Markov chain Monte Carlo.  Variational inference has been used to
approximate the posterior in many Bayesian nonparametric
models~\citep{Kurihara:2007, Doshi-Velez:2009b, Paisley:2009,
  Wang:2012a} and has been of general interest in statistics
\citep{Braun:2010, Faes:2011, Ormerod:2012}.  Here we develop a
variational algorithm for correlated nonparametric Poisson matrix
factorization (Section~\ref{sec:CNPF}).

Variational inference turns approximate posterior computation into
optimization.  We set up a family of distributions over the latent
variables $\mathcal{Q} = \{q(\cdot)\}$ and then find the member that
minimizes the KL divergence to the exact posterior.  Minimizing the KL
divergence to the posterior is equivalent to maximizing the Evidence
Lower BOund (ELBO),
\begin{align}
  q^*(\cdot) = \arg \max_{q \in \mathcal{Q}} \E_{q_{\lambda}(\xi)}[\log p(y, \xi) - \log q(\xi)],
  \label{eq:elbo}
\end{align}
where $\xi$ are the latent variables and $y$ are the observations.  In
this paper we work with the mean-field family, where the approximating
distribution fully factorizes.  Each latent variable is independently
governed by its own variational parameter.

To develop a variational method for CNPF, we give a constructive
definition of the gamma process and introduce auxiliary variables for
the Gaussian process.  We then define the corresponding mean-field
family and show how to optimize the corresponding ELBO.

\paragraph{Additional latent variables.}  We first give a constructive
definition of a homogeneous gamma process.  We scale the stick
breaking construction of \citet{Sethuraman:1994} as used
in~\citet{Gopalan:2014, Zhou:2015}.  We define stick lengths $v_k$
from a beta distribution and a scaling $s$ from a gamma distribution.
The weights of the gamma process $w_k$ are from the following process,
\begin{align*}
  s & \sim \mathrm{Gamma}(\alpha, c) \\
  v_k &\sim \mathrm{Beta}(1, \alpha)\\
  w_k &= s \left(v_i \textstyle \prod_{j=1}^{k-1} (1 - v_j)\right).
\end{align*}
We treat the gamma shape $\alpha$ and rate $c$ as latent variables
(with gamma priors).

We adapt the auxiliary variable representation of zero-mean Gaussian
processes with linear kernels~\citep{Paisley:2012b} to more general
Gaussian processes.  Suppose $G_n$ is a Gaussian process with mean
$\mu_n$ and a linear kernel.  Let $d$ be a standard Gaussian vector
with same dimension as $\ell_k$.  We can write the process as
\begin{align*}
  G(\ell_k)  \edist \ell_k^\top d + \mu_n(\ell_k).
\end{align*}
This lets us evaluate likelihoods without matrix inversion.

\paragraph{The mean-field family.} With the latent variables for the
gamma and Gaussian processes in hand, we now define the mean-field
variational distribution.  We use the following approximating family
for each latent variable
\begin{align*}
  q(x_{ku}) &= \mathrm{Gamma}(\alpha_{ku}^x, \beta_{ku}^x) \\
  q(a_{ki}) &= \mathrm{Gamma}(\alpha_{ki}^a, \beta_{ki}^a) \\
  q(s) q(V_k) q(\ell_k) &=  \delta_{\hat{s}} \delta_{\hat{V_k}} \delta_{\hat{\ell_k}} \\
  q(d_u) q(\mu_u) &= \delta_{\hat{d_u}} \delta_{\hat{\mu_u}}  \\
  q(\alpha) q(c) &= \delta_{\hat{\alpha}} \delta_{\hat{c}},
\end{align*}
where $\delta_r$
represents a point mass at $r$. As in prior work on variational
inference for Bayesian nonparametrics, we use delta distributions in
the top level stick components, scaling, and hyperparameters for
analytic
tractability~\citep{Liang:2007,Paisley:2012b,Gopalan:2014}.\footnote{This
  corresponds to variational expectation-maximization, where the E
  step computes variational expectations and the M step takes MAP
  estimates of the latent variables with delta
  factors~\citep{Beal:2003}.}

Bayesian nonparametric models contain an infinite number of latent
variables.  Following~\citet{Blei:2005}, we truncate the variational
approximation of the sticks $V_k$ and associated tuples to $T$.  In
practice it is straightforward to recognize if the truncation level is
too small because all of the components will be populated in the
fitted variational distribution.  In our studies $T=200$ was
sufficient (Section~\ref{sec:experiments}).

The goal of variational inference is to find the variational
parameters---the free parameters of $q$, such as $\hat{\ell}_k$---that
maximize the evidence lower bound.  In \myappendix{inference}, we
describe how to optimize the ELBO (Equation~\ref{eq:elbo}) with
stochastic variational inference~\citep{Hoffman:2013}. Code will be 
made available on GitHub.

We have derived a variational inference algorithm for one example of a
correlated random measure model.  Deriving algorithms for other
examples follows a similar recipe.  In general, we can handle
inference for covariance functions with inducing
variables~\citep{Titsias:2009} and subsampling~\citep{Hensman:2013}.
Further, we can address models with intractable expectations---e.g.,
those arising from different transformation distributions or Levy
measures---with recent methods for generic and nonconjugate
variational inference~\citep{Salimans:2013, Ranganath:2014,Wang:2012}.

\newcommand{\yobs}{y_{\textrm{obs}}}
\newcommand{\ytest}{y_{\textrm{test}}}

\section{Empirical Study}
\label{sec:experiments}
We study correlated nonparametric Poisson factorization (CNPF)
and compare to its uncorrelated counterpart on a large text data set and
a large data set of medical diagnosis codes.  Quantitatively, we find
that the correlated model gives better predictive performance.  We
also find that it reveals interesting visualizations of the posterior
components and their relationships.

\subsection{Study Details}

Before giving the results, we describe the baseline models, the
evaluation metric, and the hyperparameter settings.

\parhead{Baseline models.}  As a baseline, we compare against the
uncorrelated variant of Bayesian nonparametric Poisson factorization.
As we mentioned in Section~\ref{sec:crm}, uncorrelated random measures can be
cast in the correlated random measure framework by setting a
transformation distribution that does not depend on the Gaussian
process.

Recall that $x_{ik}$ is the weight for data point $i$ on component
$k$.  In the simplest Bayesian nonparametric Poisson factorization
model, the transformation distribution is
\begin{align*}
  x_{ik} \sim \textrm{Gamma}(w_k, 1).
\end{align*}

This is a two-layer hierarchical gamma process, and we abbreviate this
model HGP.  The first layer contains shared atoms and weights. The
second layer is a gamma process for each data point (e.g., patient or
document), with base measure given by the first layer's measure.

The second uncorrelated model places further hierarchy on the log of the scale
parameter of the Gamma,
\begin{align*}
  x_{ik} \sim \textrm{Gamma}\left(w_k, \exp(-m_i) \right).
\end{align*}
Here $m_i \sim \textrm{Normal}(a, b)$, which captures variants in the
row sums for each data point (i.e., how many total diagnoses for a
patient or how many words for a document). We call this model the
scaled HGP.

\myappendix{svi-baseline} gives inference details for both uncorrelated models.

\parhead{Evaluation metric.}  We compare models with held out
perplexity, a standard metric from information retrieval that relates
to held out predictive likelihood~\citep{Geisser:1975}.  We use the partial
observation scenario that is now common in topic
modeling~\citep{Wallach:2009a}.  The idea is to uncover components
from most of the data, and then evaluate how well those components can
help predict held out portions of new partially-observed data.

For each data set, we hold out 1,000 examples (i.e., rows of the
matrix).  From the remaining examples we run approximate posterior
inference, resulting in approximate posterior components $\E[a_k]$
that describe the data.  With the 1,000 held out examples, we then
split each observation (i.e., columns) randomly into two parts, 90\%
in one part ($\ytest$) and 10\% in the other ($\yobs$).  We condition
on the $\yobs$ (and that there is a test word) and calculate the
conditional perplexity on $\ytest$.  A better model will assign the
true observations a higher probability and thus lower perplexity.
Formally, perplexity is defined as
\begin{align*}
\textrm{Perplexity} = \exp{\left(\frac{-\sum_{y \in \textrm{held out}} \sum_{w \in \ytest} \log p(w \g \yobs)}{N_\textrm{held out words}} \right)}.
\end{align*}
Perplexity measures the average surprise of the test observations. The exponent is the
average number of nats (base $e$ bits) needed to encode the test sample.

\begin{landscape}
  \begin{figure*}
    \vskip -.5in
    \centering
    \includegraphics[width=1.45 \textheight]{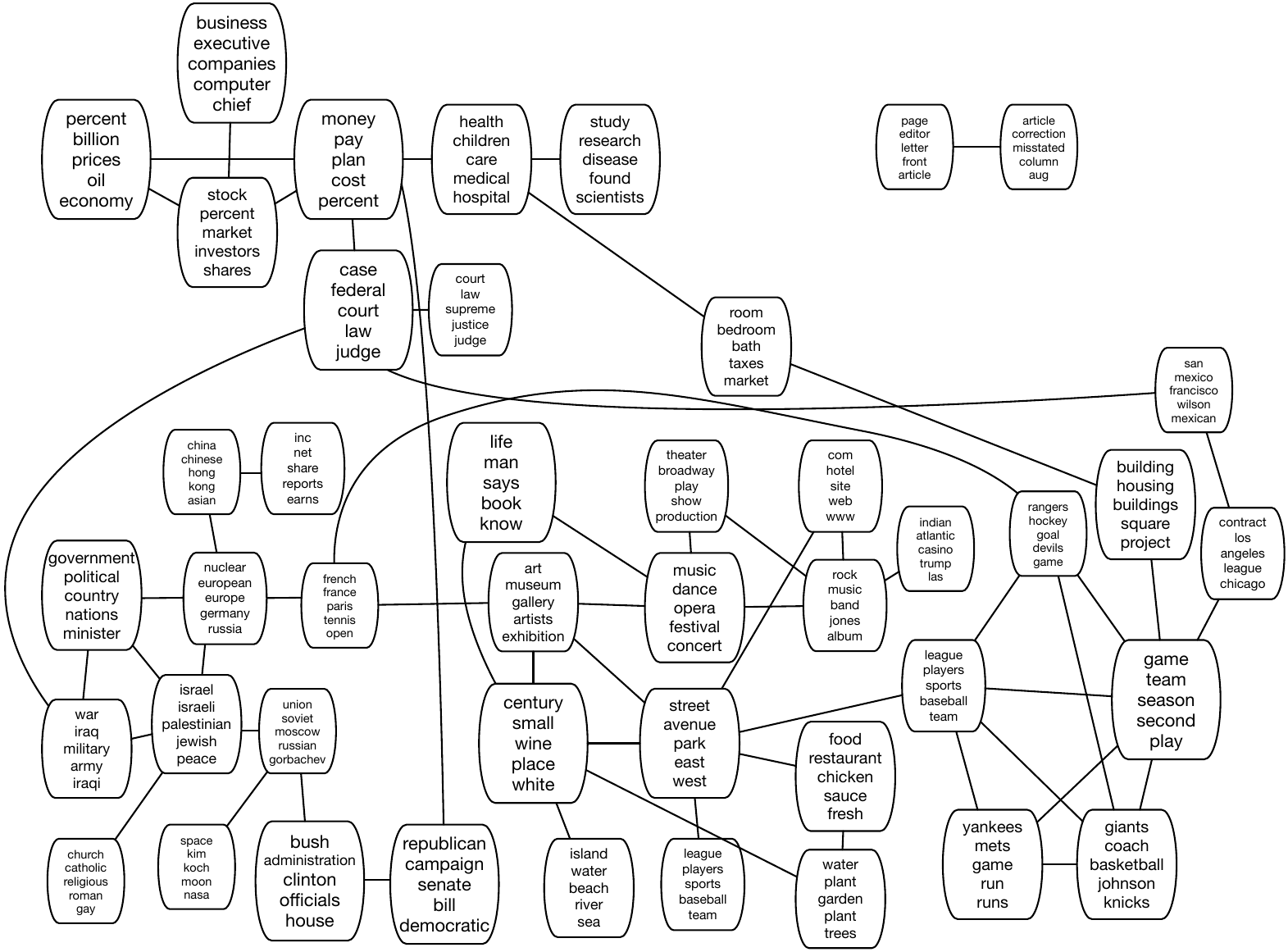}
    \caption{A graph of the latent component correlations found on
      \emph{The New York Times}.  This figure is best viewed on a
      computer with zoom.  The sizes of the components are related to
      their frequency. The correlation structure consists of several
      tightly connected groups with sparse links between them.}
    \label{fig:nyt-corr}
  \end{figure*}
\end{landscape}

\begin{landscape}
\begin{table}[tb]
\centering
\hspace*{-.6cm}
\begin{tabular}{  l  }
\toprule
1: Mammogram, Routine medical exam, Lumbago, Cervical Cancer Screening, Hypothyroidism \\
2: Hypertension, Hyperlipidemia, Coronary atherosclerosis, Prostate Cancer Screening, Vaccine for influenza \\
3: Acute pharyngitis, Cough, Myopia, Vaccine for influenza, Joint pain-shlder\\
4: Child Exam, Vaccine for flu, Otitis media, Upper respiratory infection, pharyngitis\\
5: Long-term anticoagulants, Atrial fibrillation, Hypertension, Congestive Heart Failure, Chronic airway obstruction \\
6: Normal pregnancy, Normal first pregnancy, Cervical cancer screening, Delivery, Conditions antepartum \\
7: Diabetes-2, Hypertension, Hyperlipidemia, Uncontrolled Diabetes-2, Diabetes-2 with ophthalmic manifestations  \\
8: Depression, Dysthymia, Anxiety state, Generalized anxiety disorder, Major depressive affective disorder \\
9: Joint pain lower leg, Arthritis lower leg, Local arthritis lower leg, Post-procedural status, Follow-up surgery \\
10: Allergic rhinitis, Desensitization to allergens, Asthma, Chronic allergic conjunctivitis, Chronic sinusitis  \\
11: Heart valve replacement, Prostate cancer, Lung and bronchus cancer, Secondary bone cancer, Other lung disease \\
12: Morbid obesity, Obesity, Obstructive sleep apnea, Sleep apnea, Intestinal bypass status \\
13: Acne, Convulsions, Abnormal involuntary movements, Cerebral palsy, Long-term use meds  \\
14: Abnormality of gait, Personality change, Persistent mental disorders, Lack of coordination, Debility  \\
15: Attention disorder w hyperactivity, Attention disorder w/o hyperactivity,  Adjustment
disorder, Opposition defiant disorder, Conduct disturbance  \\
16: Diseases of nail, Corns and callosities, Dermatophytosis of nail, Ingrowing nail, 
 Other states following surgery of eye and adnexa \\
17: Alcohol dependence, Tobacco use disorder, Alcohol abuse, Other alcohol dependence-in remission, 
Other alcohol dependence-continuous \\
18:  Schizophrenia-Paranoid, Long-term use meds, Schizophrenia, Schizophrenia-paranoid-chronic, Drug monitor \\
19: Female breast cancer, Personal history of breast cancer, Lymph cancer, Carcinoma in situ of breast, lymphedema \\
20: Child health exam, Vaccination for disease, Vaccinations against pneumonia, 
Need for prophylactic vaccination against viral hepatitis, Procedure \\
\bottomrule
\end{tabular}
\hspace*{0cm}
\caption{
The top twenty components on the Mayo Clinic data. We find that each factor forms a medically meaningful grouping of diagnosis codes.
For example, there are allergy, pregnancy, and alcohol dependence components.
}  \label{tab:mayo-topics}
\vskip -0.10in
\end{table}
\end{landscape}

For the models we analyze, we compute this metric as follows.  For
each held out data point we hold the components fixed (i.e.,
$\E_q[a_k]$) and use the 10\% of observed columns to form a
variational expectation of the per-data point weights $\E_q[x_{ik}]$.
In all models, we compute the held out probability of unobserved
columns by using the multinomial conditioning property of Poissons.
Conditional on there being a test observation, it is assigned to a
particular column (e.g., a word or a diagnostic code) with probability
equal to that column's normalized Poisson rates. Formally,
\begin{align*}
  p(y_i = j) =
  \frac{\sum_k \E_q[x_{ik}] \E_q[a_{kj}]}{\sum_j \sum_k \E_q[x_{ik}] \E_q[a_{kj}]}.
\end{align*}
We measure the probability of the $\ytest$ columns under this
distribution. This evaluates how well the discovered components can
form predictions in new and partially observed observations.

\parhead{Hyperparameters.}  We set the hyperparameters on the base
distribution to have shape $.01$ and rate $10.0$. We set the
truncation level $T$ to be 200, and found that none of the studies
required more than this.  We set the dimensionality of the latent
locations to be 25 and the prior variance to be $\frac{1}{250}$. We
keep these hyperparameters fixed for all data.

In the algorithm, we use Robbins Monro learning rates,
$(50 + t)^{-.9}$ for the text data and $(100 + t)^{-.9}$ for the
medical codes, and click data.

\subsection{Results}

We evaluate our posterior fits on text, medical diagnosis data, and click data.

\parhead{\emph{The New York Times}.} We study a large collection of
text from \textit{The New York Times}.  Rows are documents; columns
are vocabulary words; the cell $y_{ij}$ is the number of times term
$j$ appeared in document $i$.  After preprocessing, the data contains
100,000 documents over a vocabulary of 8,000 words.  Analyzing text
data with a Poisson factorization model is a type of topic
modeling~\citep{Blei:2012}.

Table~\ref{tab:res} summarizes the held-out perplexity. We find that
the correlated model outperforms both of the uncorrelated models. Note
that even in the uncorrelated model, adding a per-document scale
parameter improves predictions.

\begin{table}[tb]
\centering
\begin{tabular}{cccccc}
  \toprule
  & Data & HGP & Scaled HGP & CNPF &  Softplus-CNPF  \\
  \midrule
  & NYT & 3570 & 3283 & 2755 & 2768 \\
  & Mayo Clinic & 1251 & 877 & 779 & 780 \\
  & ArXiv & 5713 & 4076 & 2107 & 2120 \\
  \bottomrule
\end{tabular}
\caption{
  A summary of the predictive results on \emph{The New York Times}, the Mayo Clinic
  and ArXiv clicks. The correlated models outperform both the uncorrelated models. 
  Adding per observation scalings improves predictions.
}  \label{tab:res}
\vskip -0.10in
\end{table}

\begin{table}[tb]
\centering
\hspace*{-2.5cm}
\begin{tabular}{ ccc }
\toprule
&israel, israeli, palestinian, jewish, peace &\\
&league, players, sports, baseball, team &\\
\midrule
&room, bedroom, bath, taxes, market &\\
&news, book, magazine, editor, books &\\
\midrule
&war, iraq, military, army, iraqi &\\
&space, kim, koch, moon, nasa &\\
\midrule
&rock, music, band, jones, album &\\
&family, tax, board, paid, friend &\\
\midrule
&water, plant, garden, plants, trees &\\
&union, soviet, moscow, russian, gorbachev &\\
\midrule
&island, water, beach, river, sea &\\
&theater, broadway, play, show, production &\\
\midrule
&building, housing, buildings, square, project &\\
&news, book, magazine, editor, books &\\
\midrule
&bush, administration, clinton, officials, house &\\
&space, kim, koch, moon, nasa &\\
\midrule
&room, bedroom, bath, taxes, market &\\
&indian, atlantic, casino, trump, las &\\
\midrule
&century, small, wine, place, white & \\
&contract, los, angeles, league, chicago & \\
\bottomrule
\end{tabular}
\hspace*{-2.5cm}
\caption{
The top ten pairs of negatively correlated components inferred from \emph{the New York Times}. 
Each pair of components are highly unlikely to cooccur in an article.
}  \label{tab:nyt-neg-corr}
\vskip -0.10in
\end{table}

\begin{figure*}
  \centering
      \includegraphics[width=\textwidth]{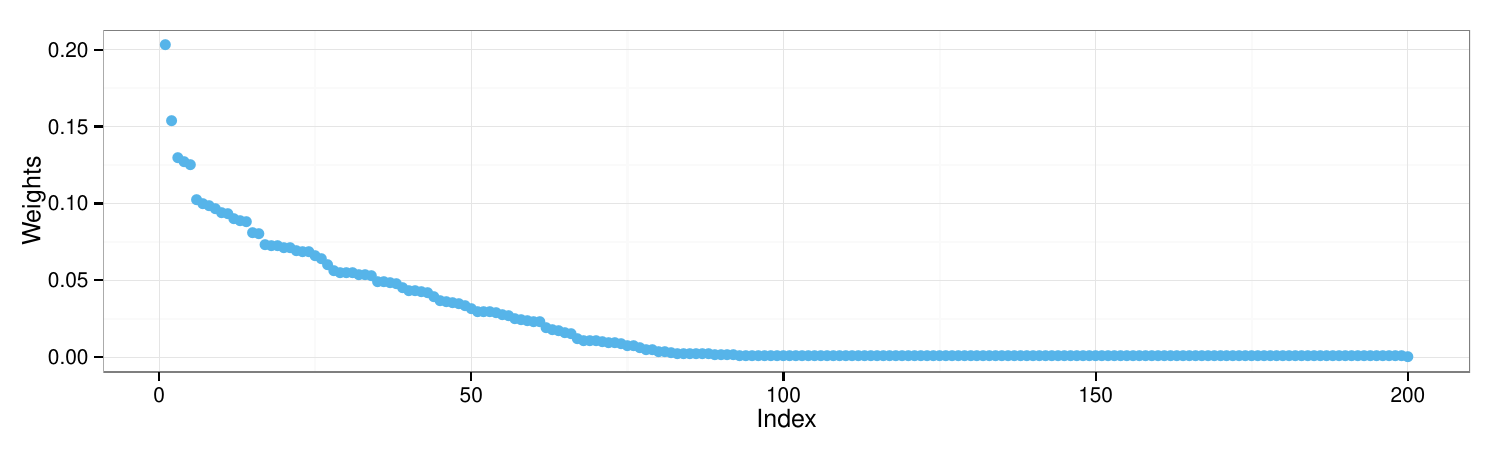}
  \caption{The weights in each tuple on  \emph{the New York Times} ordered by magnitude. Around 75 components are used.}
  \label{fig:nyt-sticks}
\end{figure*}

The model also provides new ways to explore and summarize the data.
Figure~\ref{fig:nyt-corr} is a graph of the positive correlation structure
in the posterior for the top fifty components, sorted by frequency;
Table~\ref{tab:nyt-neg-corr} contains a list of the top ten negative
correlations.  To explore the data, we compute correlations between
components by using their latent locations through the covariance
function of the Gaussian process. For these fits, the covariance
between $\ell_i$ and $\ell_j$ is $\ell_i^\top \ell_j$; the correlation
between two components is thus
\begin{align*}
  \rho_{km} = \frac{\ell_k^\top \ell_m }{\sqrt{\ell_k^\top \ell_k \ \ell_m^\top
  \ell_m}}.
\end{align*}
We find the correlation structures contains highly connected
groups, connected to each other by ``glue'', individual components
that bridge larger groups.
For example the bottom
left connected group of ``international politics" is glued together
with the top left group of ``finance" through the ``political parties"
component and the ``law'' component.

As we said above, we set the truncation level of the approximate
posterior to 200 components. Figure~\ref{fig:nyt-sticks} plots the
atom weights of these 200 components, ordered by size.  The posterior
uses about 75 components; the truncation level is appropriate.

\begin{figure*}
  \centering
      \includegraphics[width=\textwidth]{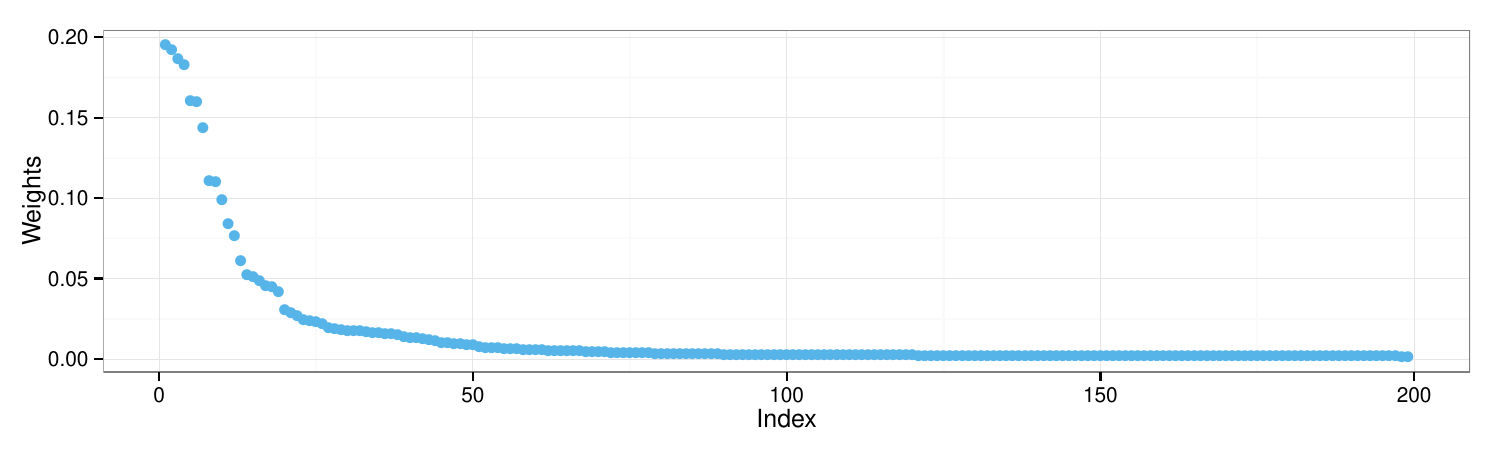}
  \caption{The weights in each tuple ordered by magnitude. Around 50 of components are used.
                      Though similar in size to the NYT data set, fewer components are used. The component
         usage has a steeper decline.}
  \label{fig:mayo-sticks}
\end{figure*}

\parhead{Medical history from the Mayo Clinic.}  We study medical code
data from the Mayo Clinic. This data set contains of all the
International Classification of Diseases 9 (ICD-9) diagnosis codes
(also called billing codes) for a collection of patients over three
years.  The diagnosis codes mark medical conditions, such as chronic
ischemic heart disease, pure hypercholesterolemia, and type 2
diabetes.  The entire collection contains 142,297 patients and 11,102
codes.  Patients are rows in the matrix; codes are columns; each cell
marks how many times the patient was assigned to the code.

Table~\ref{tab:res} summarizes the held-out perplexity.  Again, the
correlated model does best. Further, as for text modeling, it is
important to allow a patient-specific scale parameter to capture their
relative health.  Figure~\ref{fig:mayo-sticks} plots the posterior
sticks, ordered by size.  The approximate posterior uses about 50
components, using the first 20 more heavily.

Table~\ref{tab:mayo-topics} contains the 20 most commonly used
components. The components correspond to medically meaningful groups
of conditions, such as obesity (12), type 2 diabetes (7), and breast
malignancy (19).  The top positive correlations are in
Table~\ref{tab:mayo-corr}. There are several meaningful correlations, such
as depression \& alcohol dependency, and using anticoagulants \&
hypertension/lipidemia.  Note that the relationship between
schizophrenia and type 2 diabetes is an active area of research in
medicine \citep{Med-Suvisaari:2008, Med-Liu:2013}.

\begin{table}[tb]
\centering
\hspace*{-1.75cm}
\begin{tabular}{ lll }
\toprule
-- Long-term anticoagulants, Atrial fibrillation, Hypertension, Congestive Heart Failure, Chronic airway obstruction \\
-- Heart valve replacement, Prostate cancer, Lung and bronchus cancer, Secondary bone cancer, Other lung disease \\
\midrule
-- Abnormality of gait, Personality change, Persistent mental disorders, Lack of coordination, Debility \\
-- Schizophrenia-Paranoid, Long-term use meds, Schizophrenia, Schizophrenia-paranoid-chronic, Drug monitor \\
\midrule
-- Attention deficit disorder with hyperactivity, Attention deficit disorder without hyperactivity,  Adjustment
\\ \hspace{1cm}  disorder with disturbance of emotions and conduct, Opposition defiant disorder, Conduct disturbance  \\
-- Schizophrenia-Paranoid, Long-term use meds, Schizophrenia, Schizophrenia-paranoid-chronic, Drug monitor \\
\midrule
-- Depression, Dysthymia, Anxiety state, Generalized anxiety disorder, Major depressive affective disorder  \\
-- Alcohol dependence, Tobacco use disorder, Alcohol abuse, Other alcohol dependence-in remission, 
\\ \hspace{1cm} Other alcohol dependence-continuous \\
\midrule
-- Long-term anticoagulants, Atrial fibrillation, Hypertension, Congestive Heart Failure, Chronic airway obstruction \\
-- Diabetes-2, Hypertension, Hyperlipidemia, Uncontrolled Diabetes-2, Diabetes-2 with ophthalmic manifestations  \\
\midrule
-- Hypertension, Hyperlipidemia, Coronary atherosclerosis, Prostate Cancer Screening, Vaccine for influenza \\
-- Long-term anticoagulants, Atrial fibrillation, Hypertension, Congestive Heart Failure, Chronic airway obstruction \\
\midrule
-- Heart valve replacement, Prostate cancer, Lung and bronchus cancer, Secondary bone cancer, Other lung disease \\
-- Female breast cancer, Personal history of breast cancer, Lymph cancer, Carcinoma in situ of breast, Lymphedema  \\
\midrule
-- Depression, Dysthymia, Anxiety state, Generalized anxiety disorder, Major depressive affective disorder  \\
-- Schizophrenia-Paranoid, Long-term use meds, Schizophrenia, Schizophrenia-paranoid-chronic, Drug monitor \\
\midrule
-- Diabetes-2, Hypertension, Hyperlipidemia, Uncontrolled Diabetes-2, Diabetes-2 with ophthalmic manifestations  \\
-- Schizophrenia-Paranoid, Long-term use meds, Schizophrenia, Schizophrenia-paranoid-chronic, Drug monitor \\
\midrule
-- Mammogram, Routine medical exam, Lumbago, Cervical Cancer Screening, Hypothyroidism \\
-- Female breast cancer, Personal history of breast cancer, Lymph cancer, Carcinoma in situ of breast, Lymphedema \\
\bottomrule
\end{tabular}
\hspace*{-2.5cm}
\caption{
The top ten correlations among the heavily used components in the Mayo Clinic data. 
We find several medically meaningful relationships between latent components. 
For example, the relationships between obesity and type 2 diabetes is well established.
}  \label{tab:mayo-corr}
\vskip -0.10in
\end{table}

\parhead{ArXiv click data.}  Finally, we examine user click data from
the ArXiv, an online repository of research articles.  The ArXiv
initially focused on physics articles but has now expanded to many
other domains, including statistics.  This data set contains the
number of times each user clicked on an article; it spans 50,000 users
and 20,000 articles.  Building models of such data is useful, for
example, to develop recommendation systems that find interesting
articles to ArXiv readers.

As for the other data, we hold out some of the clicks and try to
predict them.  Table~\ref{tab:res} summarizes the results. We find
similar results as on our other two data sets. The correlated models
outperform the uncorrelated models on predicting unseen clicks for new
users. We find that the standard CNPF model outperforms the softplus
CNPF on all of our data sets.

\section{Discussion}

We present correlated random measures.  Correlated random measures
enable us to construct versions of Bayesian nonparametric models that
capture correlations between their components. We construct several
examples of such models, and develop the inference algorithm in detail
for one of them, correlated nonparametric Poisson factorization.  With
this model, we find that the correlated random measure improves
predictions and produces interesting interpretable results.

Random probability measures such as the Dirichlet process are
consistent for density estimation, so why might one prefer a
correlated random measure over a completely random measure? We
conjecture that correlated random measures make more efficient use of
the data. One promising avenue of future research is to study the
rates of correlated random measures versus completely random measures.

Correlated random measures model latent correlations in the data,
while dependent random measures model correlations based on observed
covariates.  Combining these two ideas to incorporate correlations
both observed and latent yields a broader class of random measures
that can model many real world phenomena. Another avenue of future
research is to study this construction both methodologically and
practically.

We define correlated random measures by combining Poisson and Gaussian
processes.  However, we note that other processes can also be used for
the source of tuples $(a_i, w_i, \ell_i)$ and the random function.
For example, the DILN model of Section~\ref{sec:DILN} uses a Dirichlet
process to form its tuples; another way to generate tuples would be
through the Pitman-Yor process~\citep{Pitman:1997, Teh:2006}.

Similarly, though we used a Gaussian process to define a random
function from latent locations to real values, there are other
possibilities.  For example we can replace the GP with the student-T
process~\citep{Shah:2014}.  Or, if we restrict the latent locations to be
positive then we can use them to subordinate, as an index to, another
stochastic process, such as Brownian motion.  Also, we could use
discrete random functions to form feature groups. We leave these
extensions for possible future research.

\parhead{Acknowledgements.} 
This work is supported by NSF IIS-1247664,  ONR
N00014-11-1-0651, DARPA FA8750-14-2-0009, DARPA
N66001-15-C-4032, Adobe, NDSEG Fellowship, Porter Ogden Jacobus Fellowship,
Seibel Foundation, and the Sloan Foundation. The authors would
like to thank the reviewers for their helpful feedback and comments.

\bibliography{../bib.bib}

\begin{thebibliography}{}

\bibitem[Ammann et~al., 1978]{Ammann:1978}
Ammann, L.~P., Thall, P.~F., et~al. (1978).
\newblock Random measures with aftereffects.
\newblock {\em The Annals of Probability}, 6(2):216--230.

\bibitem[Antoniak, 1974]{Antoniak:1974}
Antoniak, C. (1974).
\newblock Mixtures of {D}irichlet processes with applications to {B}ayesian
  nonparametric problems.
\newblock {\em The Annals of Statistics}, 2(6):1152--1174.

\bibitem[Beal, 2003]{Beal:2003}
Beal, M. (2003).
\newblock {\em Variational algorithms for approximate {B}ayesian inference}.
\newblock PhD thesis, Gatsby Computational Neuroscience Unit, University
  College London.

\bibitem[Bishop, 2006]{Bishop:2006}
Bishop, C. (2006).
\newblock {\em Pattern Recognition and Machine Learning}.
\newblock Springer New York.

\bibitem[Blei, 2012]{Blei:2012}
Blei, D. (2012).
\newblock Probabilistic topic models.
\newblock {\em Communications of the ACM}, 55(4):77--84.

\bibitem[Blei and Jordan, 2005]{Blei:2005}
Blei, D. and Jordan, M. (2005).
\newblock Variational inference for {D}irichlet process mixtures.
\newblock {\em Journal of Bayesian Analysis}, 1(1):121--144.

\bibitem[Borodin, 2009]{Borodin:2009}
Borodin, A. (2009).
\newblock Determinantal point processes.
\newblock {\em arXiv preprint arXiv:0911.1153}.

\bibitem[Braun and McAuliffe, 2007]{Braun:2010}
Braun, M. and McAuliffe, J. (2007).
\newblock Variational inference for large-scale models of discrete choice.
\newblock {\em Journal of American Statistical Association}, 105(489).

\bibitem[Canny, 2004]{Canny:2004}
Canny, J. (2004).
\newblock {G}a{P}: {A} factor model for discrete data.
\newblock In {\em Proceedings of the 27th Annual International ACM SIGIR
  Conference on Research and Development in Information Retrieval}.

\bibitem[Cinlar, 2011]{Cinlar:2011}
Cinlar, E. (2011).
\newblock {\em Probability and Stochastics}.
\newblock Springer.

\bibitem[Cox and Isham, 1980]{Cox:1980}
Cox, D.~R. and Isham, V. (1980).
\newblock {\em Point Processes}.
\newblock Chapman Hall.

\bibitem[DeYoreo and Kottas, 2015]{Deyoreo:2015}
DeYoreo, M. and Kottas, A. (2015).
\newblock Modeling for dynamic ordinal regression relationships: An application
  to estimating maturity of rockfish in california.
\newblock {\em arXiv preprint arXiv:1507.01242}.

\bibitem[Doshi-Velez and Ghahramani, 2009]{Doshi-Velez:2009c}
Doshi-Velez, F. and Ghahramani, Z. (2009).
\newblock Correlated non-parametric latent feature models.
\newblock In {\em Proceedings of the Twenty-Fifth Conference on Uncertainty in
  Artificial Intelligence}, pages 143--150. AUAI Press.

\bibitem[Doshi-Velez et~al., 2009]{Doshi-Velez:2009b}
Doshi-Velez, F., Miller, K., Van~Gael, J., and Teh, Y. (2009).
\newblock Variational inference for the {I}ndian buffet process.
\newblock In {\em Proceedings of the Intl. Conf. on Artificial Intelligence and
  Statistics}, pages 137--144.

\bibitem[Dunson and Herring, 2005]{Dunson:2005}
Dunson, D.~B. and Herring, A.~H. (2005).
\newblock Bayesian latent variable models for mixed discrete outcomes.
\newblock {\em Biostatistics}, 6(1):11--25.

\bibitem[Escobar and West, 1995]{Escobar:1995}
Escobar, M. and West, M. (1995).
\newblock Bayesian density estimation and inference using mixtures.
\newblock {\em Journal of the American Statistical Association}, 90:577--588.

\bibitem[Faes et~al., 2011]{Faes:2011}
Faes, C., Ormerod, J.~T., and Wand, M.~P. (2011).
\newblock Variational bayesian inference for parametric and nonparametric
  regression with missing data.
\newblock {\em Journal of the American Statistical Association}, 106(495).

\bibitem[Ferguson, 1973]{Ferguson:1973}
Ferguson, T. (1973).
\newblock A {B}ayesian analysis of some nonparametric problems.
\newblock {\em The Annals of Statistics}, 1:209--230.

\bibitem[Foti et~al., 2013]{Foti:2013}
Foti, N.~J., Futoma, J.~D., Rockmore, D.~N., and Williamson, S. (2013).
\newblock A unifying representation for a class of dependent random measures.
\newblock In {\em International Conference on Artifical Intelligence and
  Statistics}.

\bibitem[Foti et~al., 2015]{Foti:2015}
Foti, N.~J., Williamson, S., et~al. (2015).
\newblock A survey of non-exchangeable priors for bayesian nonparametric
  models.
\newblock {\em Pattern Analysis and Machine Intelligence, IEEE Transactions
  on}, 37(2):359--371.

\bibitem[Fox et~al., 2011]{Fox:2011}
Fox, E.~B., Sudderth, E.~B., Jordan, M.~I., and Willsky, A.~S. (2011).
\newblock A sticky hdp-hmm with application to speaker diarization.
\newblock {\em The Annals of Applied Statistics}, pages 1020--1056.

\bibitem[Gasthaus et~al., 2009]{Gasthaus:2009}
Gasthaus, J., Wood, F., Gorur, D., and Teh, Y.~W. (2009).
\newblock Dependent dirichlet process spike sorting.
\newblock In {\em Advances in neural information processing systems}, pages
  497--504.

\bibitem[Geisser, 1975]{Geisser:1975}
Geisser, S. (1975).
\newblock The predictive sample reuse method with applications.
\newblock {\em Journal of the American Statistical Association},
  70(350):320--328.

\bibitem[Ghahramani, 1995]{Ghahramani:1995}
Ghahramani, Z. (1995).
\newblock Factorial learning and the em algorithm.
\newblock In {\em Advances in Neural Information Processing Systems}, pages
  617--624.

\bibitem[Ghahramani and Jordan, 1997]{Ghahramani:1997}
Ghahramani, Z. and Jordan, M. (1997).
\newblock Factorial hidden {M}arkov models.
\newblock {\em Machine Learning}, 31(1).

\bibitem[Gopalan et~al., 2014]{Gopalan:2014}
Gopalan, P., Ruiz, F.~J., Ranganath, R., and Blei, D.~M. (2014).
\newblock Bayesian nonparametric poisson factorization for recommendation
  systems.
\newblock In {\em International Conference on Artificial Intelligence and
  Statistics}.

\bibitem[Griffin and Leisen, 2014]{Griffin:2014}
Griffin, J.~E. and Leisen, F. (2014).
\newblock Compound random measures and their use in bayesian nonparametrics.
\newblock {\em arXiv preprint arXiv:1410.0611}.

\bibitem[Griffiths and Ghahramani, 2006]{Griffiths:2006a}
Griffiths, T. and Ghahramani, Z. (2006).
\newblock Infinite latent feature models and the {I}ndian buffet process.
\newblock In {\em Advances in Neural Information Processing Systems (NIPS)}.

\bibitem[Hensman et~al., 2013]{Hensman:2013}
Hensman, J., Fusi, N., and Lawrence, N.~D. (2013).
\newblock Gaussian processes for big data.
\newblock In {\em Conference on Uncertainty in Artificial Intelligence}.

\bibitem[Hjort, 1990]{Hjort:1990}
Hjort, N. (1990).
\newblock Nonparametric bayes estimators based on beta processes in models for
  life history data.
\newblock {\em The Annals of Statistics}, 18(3).

\bibitem[Hoffman et~al., 2013]{Hoffman:2013}
Hoffman, M., Blei, D., Wang, C., and Paisley, J. (2013).
\newblock Stochastic variational inference.
\newblock {\em Journal of Machine Learning Research}, 14(1303--1347).

\bibitem[Honkela et~al., 2008]{Honkela:2008}
Honkela, A., Tornio, M., Raiko, T., and Karhunen, J. (2008).
\newblock Natural conjugate gradient in variational inference.
\newblock In {\em Neural Information Processing}.

\bibitem[Ishwaran and Rao, 2005]{Ishwaran:2005}
Ishwaran, H. and Rao, J.~S. (2005).
\newblock Spike and slab variable selection: {F}requentist and {B}ayesian
  strategies.
\newblock {\em The Annals of Statistics}, 33(2):730--773.

\bibitem[Jordan et~al., 1999]{Jordan:1999}
Jordan, M., Ghahramani, Z., Jaakkola, T., and Saul, L. (1999).
\newblock Introduction to variational methods for graphical models.
\newblock {\em Machine Learning}, 37:183--233.

\bibitem[Kingman, 1967]{Kingman:1967}
Kingman, J. (1967).
\newblock Completely random measures.
\newblock {\em Pacific Journal of Mathematics}, 21(1).

\bibitem[Kingman, 1993]{Kingman:1993}
Kingman, J. (1993).
\newblock {\em Poisson Processes}.
\newblock Oxford University Press, USA.

\bibitem[Kurihara et~al., 2007]{Kurihara:2007}
Kurihara, K., Welling, M., and Teh, Y. (2007).
\newblock Collapsed variational {D}irichlet process mixture models.
\newblock In {\em International Joint Conferences on Artificial Intelligence
  (IJCAI)}.

\bibitem[Liang et~al., 2007]{Liang:2007}
Liang, P., Petrov, S., Klein, D., and Jordan, M. (2007).
\newblock The infinite {PCFG} using hierarchical {D}irichlet processes.
\newblock In {\em Empirical Methods in Natural Language Processing}.

\bibitem[Liu et~al., 2013]{Med-Liu:2013}
Liu, Y., Li, Z., Zhang, M., Deng, Y., Yi, Z., and Shi, T. (2013).
\newblock Exploring the pathogenetic association between schizophrenia and type
  2 diabetes mellitus diseases based on pathway analysis.
\newblock {\em BMC medical genomics}, 6(Suppl 1):S17.

\bibitem[MacEachern, 1999]{MacEachern:1999}
MacEachern, S. (1999).
\newblock Dependent nonparametric processes.
\newblock In {\em ASA Proceedings of the Section on Bayesian Statistical
  Science}.

\bibitem[Miller et~al., 2009]{Miller:2009}
Miller, K., Griffiths, T., and Jordan, M. (2009).
\newblock Nonparametric latent feature models for link prediction.
\newblock In Bengio, Y., Schuurmans, D., Lafferty, J., Williams, C. K.~I., and
  Culotta, A., editors, {\em Advances in Neural Information Processing Systems
  22}, pages 1276--1284.

\bibitem[M{\o}ller et~al., 1998]{Moller:1998}
M{\o}ller, J., Syversveen, A.~R., and Waagepetersen, R.~P. (1998).
\newblock Log gaussian cox processes.
\newblock {\em Scandinavian journal of statistics}, 25(3):451--482.

\bibitem[Ormerod and Wand, 2012]{Ormerod:2012}
Ormerod, J.~T. and Wand, M. (2012).
\newblock Gaussian variational approximate inference for generalized linear
  mixed models.
\newblock {\em Journal of Computational and Graphical Statistics}, 21(1):2--17.

\bibitem[Paisley and Carin, 2009]{Paisley:2009}
Paisley, B. and Carin, L. (2009).
\newblock Nonparametric factor analysis with beta process priors.
\newblock In {\em International Conference on Machine Learning}.

\bibitem[Paisley et~al., 2012]{Paisley:2012b}
Paisley, J., Wang, C., and Blei, D. (2012).
\newblock The discrete infinite logistic normal distribution.
\newblock {\em Bayesian Analysis}, 7(2):235--272.

\bibitem[Paisley et~al., 2015]{Paisley:2015}
Paisley, J., Wang, C., Blei, D.~M., Jordan, M., et~al. (2015).
\newblock Nested hierarchical dirichlet processes.
\newblock {\em Pattern Analysis and Machine Intelligence, IEEE Transactions
  on}, 37(2):256--270.

\bibitem[Pitman and Yor, 1997]{Pitman:1997}
Pitman, J. and Yor, M. (1997).
\newblock The two-parameter poisson-dirichlet distribution derived from a
  stable subordinator.
\newblock {\em The Annals of Probability}, pages 855--900.

\bibitem[Ranganath et~al., 2014]{Ranganath:2014}
Ranganath, R., Gerrish, S., and Blei, D. (2014).
\newblock $\{$Black Box Variational Inference$\}$.
\newblock In {\em Proceedings of the Seventeenth International Conference on
  Artificial Intelligence and Statistics}, pages 814--822.

\bibitem[Rasmussen and Williams, 2005]{Rasmussen:2005}
Rasmussen, C.~E. and Williams, C. K.~I. (2005).
\newblock {\em Gaussian Processes for Machine Learning}.
\newblock The MIT Press.

\bibitem[Robbins and Monro, 1951]{Robbins:1951}
Robbins, H. and Monro, S. (1951).
\newblock A stochastic approximation method.
\newblock {\em The Annals of Mathematical Statistics}, 22(3):pp. 400--407.

\bibitem[Salimans et~al., 2013]{Salimans:2013}
Salimans, T., Knowles, D.~A., et~al. (2013).
\newblock Fixed-form variational posterior approximation through stochastic
  linear regression.
\newblock {\em Bayesian Analysis}, 8(4):837--882.

\bibitem[Sato, 2001]{Sato:2001}
Sato, M. (2001).
\newblock Online model selection based on the variational {B}ayes.
\newblock {\em Neural Computation}, 13(7):1649--1681.

\bibitem[Sethuraman, 1994]{Sethuraman:1994}
Sethuraman, J. (1994).
\newblock A constructive definition of {D}irichlet priors.
\newblock {\em Statistica Sinica}, 4:639--650.

\bibitem[Shah et~al., 2014]{Shah:2014}
Shah, A., Wilson, A.~G., and Ghahramani, Z. (2014).
\newblock Student-t processes as alternatives to gaussian processes.
\newblock {\em arXiv preprint arXiv:1402.4306}.

\bibitem[Sudderth and Jordan, 2009]{Sudderth:2009}
Sudderth, E.~B. and Jordan, M.~I. (2009).
\newblock Shared segmentation of natural scenes using dependent pitman-yor
  processes.
\newblock In {\em Advances in Neural Information Processing Systems}, pages
  1585--1592.

\bibitem[Suvisaari et~al., 2008]{Med-Suvisaari:2008}
Suvisaari, J., Per{\"a}l{\"a}, J., Saarni, S.~I., H{\"a}rk{\"a}nen, T.,
  Pirkola, S., Joukamaa, M., Koskinen, S., L{\"o}nnqvist, J., and Reunanen, A.
  (2008).
\newblock Type 2 diabetes among persons with schizophrenia and other psychotic
  disorders in a general population survey.
\newblock {\em European Archives of Psychiatry and Clinical Neuroscience},
  258(3):129--136.

\bibitem[Teh, 2006]{Teh:2006a}
Teh, Y. (2006).
\newblock A hierarchical {B}ayesian language model based on {P}itman-{Y}or
  processes.
\newblock In {\em Proceedings of the Association of Computational Linguistics}.

\bibitem[Teh et~al., 2006]{Teh:2006}
Teh, Y.~W., Jordan, M., Beal, M.~J., and Blei, D.~M. (2006).
\newblock Hierarchical dirichlet processes.
\newblock {\em Journal of the American Statistical Association},
  101:1566--1581.

\bibitem[Teh and Jordan, 2010]{Teh:2010a}
Teh, Y.~W. and Jordan, M.~I. (2010).
\newblock Hierarchical {B}ayesian nonparametric models with applications.
\newblock In Hjort, N., Holmes, C., M{\"u}ller, P., and Walker, S., editors,
  {\em Bayesian Nonparametrics: Principles and Practice}. Cambridge University
  Press.

\bibitem[Thibaux and Jordan, 2007]{Thibaux:2007}
Thibaux, R. and Jordan, M. (2007).
\newblock Hierarchical beta processes and the {I}ndian buffet process.
\newblock In {\em 11th Conference on Artificial Intelligence and Statistics}.

\bibitem[Tieleman and Hinton, 2012]{Tieleman:2012}
Tieleman, T. and Hinton, G. (2012).
\newblock Lecture 6.5-rmsprop: Divide the gradient by a running average of its
  recent magnitude.
\newblock In {\em COURSERA: Neural Networks for Machine Learning}.

\bibitem[Titsias, 2008]{Titsias:2008}
Titsias, M.~K. (2008).
\newblock The infinite gamma-poisson feature model.
\newblock In {\em Advances in Neural Information Processing Systems}, pages
  1513--1520.

\bibitem[Titsias, 2009]{Titsias:2009}
Titsias, M.~K. (2009).
\newblock Variational learning of inducing variables in sparse gaussian
  processes.
\newblock In {\em International Conference on Artificial Intelligence and
  Statistics}, pages 567--574.

\bibitem[Wainwright and Jordan, 2008]{Wainwright:2008}
Wainwright, M. and Jordan, M. (2008).
\newblock Graphical models, exponential families, and variational inference.
\newblock {\em Foundations and Trends in Machine Learning}, 1(1--2):1--305.

\bibitem[Wallach et~al., 2009]{Wallach:2009a}
Wallach, H., Murray, I., Salakhutdinov, R., and Mimno, D. (2009).
\newblock Evaluation methods for topic models.
\newblock In {\em International Conference on Machine Learning (ICML)}.

\bibitem[Wang and Blei, 2012]{Wang:2012a}
Wang, C. and Blei, D.~M. (2012).
\newblock Truncation-free stochastic variational inference for {B}ayesian
  nonparametric models.
\newblock In {\em Advances in Neural Information Processing Systems (NIPS)}.

\bibitem[{Wang} and {Blei}, 2013]{Wang:2012}
{Wang}, C. and {Blei}, D.~M. (2013).
\newblock Variational inference for nonconjutate models.
\newblock {\em JMLR}.

\bibitem[Williamson et~al., 2010]{Williamson:2010b}
Williamson, S., Wang, C., Heller, K.~A., and Blei, D.~M. (2010).
\newblock The ibp compound dirichlet process and its application to focused
  topic modeling.
\newblock In {\em Proceedings of the 27th International Conference on Machine
  Learning (ICML-10)}, pages 1151--1158.

\bibitem[Zhou and Carin, 2015]{Zhou:2015}
Zhou, M. and Carin, L. (2015).
\newblock Negative binomial process count and mixture modeling.
\newblock {\em Pattern Analysis and Machine Intelligence}.

\bibitem[Zhou et~al., 2009]{Zhou:2009}
Zhou, M., Chen, H., Paisley, J., Ren, L., Sapiro, G., and Carin, L. (2009).
\newblock Non-parametric bayesian dictionary learning for sparse image
  representations.
\newblock In Bengio, Y., Schuurmans, D., Lafferty, J., Williams, C. K.~I., and
  Culotta, A., editors, {\em Advances in Neural Information Processing Systems
  22}, pages 2295--2303.

\bibitem[Zhou et~al., 2012]{Zhou:2012}
Zhou, M., Hannah, L., Dunson, D., and Carin, L. (2012).
\newblock Beta negative binomial process and poisson factor analysis.
\newblock In {\em International Conference on Artificial Intelligence and
  Statistics}.

\bibitem[Zhou et~al., 2011]{Zhou:2011}
Zhou, M., Yang, H., Sapiro, G., Dunson, D.~B., and Carin, L. (2011).
\newblock Dependent hierarchical beta process for image interpolation and
  denoising.
\newblock In {\em International conference on artificial intelligence and
  statistics}, pages 883--891.

\end{thebibliography}
\bibliographystyle{apalike}

 \appendix
\section{Appendix}
In the appendix we describe the Laplace transform of CorrRMs, 
establish conditions for integrability, derive inference for correlated nonparametric
Poisson factorization, 
and show the changes
to inference needed for variational inference in our comparison models. 

\subsection{Laplace Transform}
 \label{appendix:laplace-functional}
We can characterize the Laplace functional of a Poisson driven
correlated random measure in terms of a Gaussian expectation.
\begin{proposition}
Let $M$ be drawn from a correlated random measure with Poisson
mean measure $\nu$, Gaussian process parameters $m$ and $K$ with
Gaussian process draw $F$,
transformation distribution $p(x | \cdot)$, and
let $g$ be a positive, real valued, $\cE$ measurable function, then
the Laplace functional $\E[e^{Mg}] = \E_F[e^{-\tilde{\nu} (1 - e^{-r g(a) x})}]$,
where $\tilde{\nu} = \nu(da, dw, d \ell)  p(dx | F(\ell_i), w_i)$.
\label{prop:laplace_functional}
\end{proposition}
The Laplace functional can be used for analytic computation of properties of correlated random
measures as it characterizes all moments of integrals with respect to this random measure.

\paragraph{Proof of Proposition \ref{prop:laplace_functional}.}
Let $(a_i, w_i, \ell_i)_{i \in \cI}$ be the atoms of the Poisson random measure drawn with mean $\nu$. Then,
conditional on $F$ by the transformation property of Poisson random measures $(a_i, w_i, \ell_i, x_i)$ is a Poisson random
measure. This follows as given $F, w_i, \ell_i$, the $x_i$ are conditionally independent \citep{Kingman:1993}. The mean
measure of this Poisson random measure given $F$ is 
\begin{align}
\tilde{\nu} = \nu(da, dw, d \ell) p(dx | F(\ell_i), w_i).
\end{align}
The correlated random measure $M$ can be written as integral with respect to the Poisson random
measure $N$ with mean $\tilde{\nu}$ as $N f(a) x$. Thus,
\begin{align*}
\E[e^{-r M g}] = \E[\E[e^{-r M g} | F]] = \E[\E[e^{-r N g(a) x} | F]] = \E[e^{-\tilde{\nu} (1 - e^{-r g(a) x})}],
\end{align*}
where the last step follows from the Laplace functional of a Poisson process.

\subsection{Integrability}
 \label{appendix:integrability}
Establishing conditions for integrability with respect to the
random measure aids in the construction of models (consider the
aforementioned correlated probability measures). 
Here we provide a proposition (using the notation $x \wedge y$ is 
the smaller of x and y)
that completely characterizes integrability of positive functions with respect
to a Poisson driven correlated random measure.
\begin{proposition}
Let $M$ be drawn from a correlated random measure with Poisson
mean measure $\nu$, Gaussian process parameters $m$ and $K$ with
Gaussian process draw $F$, 
transformation distribution
$p(x | \cdot)$, and
let $g$ be a positive, real valued, $\cE$ measurable function. Then, $Mg$ is finite with
probability $\cP_F(\tilde{\nu} g(a)x \wedge 1 < \infty)$ ,
where $\tilde{\nu} = \nu(da, dw, d \ell)  p(dx | F(\ell_i), w_i)$.
\label{prop:integrability}
\end{proposition}
Note the probability is a Gaussian expectation. 
This proposition parallels the integrability conditions based on the mean
measure for Poisson random measures \citep{Cinlar:2011}. We use this
proposition to establish finiteness in our examples.

\paragraph{Proof of Proposition \ref{prop:integrability}.}
We first begin by noting that 
\begin{align*}
\cP(Mg < \infty) = 
\lim_{r \to 0} \E e^{-r M g} = \lim_{r \to 0} \E[e^{-\tilde{\nu} (1 - e^{-r g(a) x})}] = \E[\lim_{r \to 0} e^{-\tilde{\nu} (1 - e^{-r g(a) x})}],
\end{align*}
by Proposition \ref{prop:laplace_functional} and where the last equality follows from the dominated convergence theorem 
and the positivity of $r g(a) x$.

The function $g(a) x \wedge 1$ dominates  $(1 - e^{-r g(a) x})$ for $ r < 1$, thus when $\tilde{\nu} g(a) x \wedge 1 < \infty$,
then $\lim_{r \to 0} e^{-\tilde{\nu} (1 - e^{-r g(a) x})}=1$. Similarly $(1 - e^{-r g(a) x})$ dominates  $(1 - e^{-1})  (g(a) x \wedge 1)$, thus when 
$\tilde{\nu} g(a) x \wedge 1 = \infty$, $\lim_{r \to 0} e^{-\tilde{\nu} (1 - e^{-r g(a) x})}=0$. Putting this all together gives
\begin{align*}
\E[\lim_{r \to 0} e^{-\tilde{\nu} (1 - e^{-r g(a) x})}]  = \E[\delta(\tilde{\nu} g(a) x \wedge 1 < \infty)] = \cP_F(\tilde{\nu} g(a) x \wedge 1 < \infty).
\end{align*}
Thus $\cP(Mg < \infty) = \cP_F(\tilde{\nu} g(a) x \wedge 1 < \infty)$. 

We can use Proposition \ref{prop:integrability} finiteness of $M(E)$ by letting $g$ equal to 1 everywhere.
That is, $\cP(M(E) < \infty) = \cP_F(\tilde{\nu} x \wedge 1 < \infty)$. 

Both propositions naturally extend
to the hierarchical case when the shared tuples come from a Poisson process.

\subsection{Finiteness}
 \label{appendix:finiteness}

\paragraph{Finiteness of the CorrRM in correlated nonparametric Poisson factorization.}
The Poisson rate is an inner
product between iid gamma variables and a draw from a correlated random measure. 
Thus the Poisson rate is finite almost surely if draws from the correlated random measure
produce finite measures almost surely. We show this using Proposition \ref{prop:integrability}.

In this case the mean of the
conditional Poisson random measure is 
\begin{align*}
\tilde{\nu}(d\ell, da, dw, dx) = L(d \ell) H(da) e^{-cw}/w dw \frac{{e^{-F(\ell)}}^w}{\Gamma(w)} x^{w - 1} e^{-x e^{-F(\ell)}} dx.
\end{align*}
We seek to integrate $x \wedge 1$ with respect to this measure. Using a change of variables, this integral is equal to
\begin{align*}
\int x& \wedge 1 L(d \ell) H(da) e^{-cw}/w dw \frac{{e^{-F(\ell)}}^w}{\Gamma(w)} x^{w - 1} e^{-x e^{-F(\ell)}} dx 
\\
&\leq \int L(d \ell) H(da) e^{-cw}/w dw x \frac{{e^{-F(\ell)}}^w}{\Gamma(w)} x^{w - 1} e^{-x e^{-F(\ell)}} dx 
\\
&= H(E) \int L(d\ell) e^{-cw} e^{F(\ell)} dw
\\
&= \frac{H(E)}{c} \int e^{ F(\ell)} L(d\ell). 
\end{align*}
Thus the Poisson rate is finite when $\int e^{ F(\ell)} L(d \ell) < \infty$. By Tonelli's theorem and 
assuming $G$ has a linear kernel and a constant mean $\mu$ which has a mean zero and $\sigma_\mu^2$ variance prior,
\begin{align*}
\E[\int e^{ F(\ell)} L(d\ell)] = \int e^{\frac{1}{2} K(\ell, \ell)} L(d\ell) + \exp\left(\frac{\sigma_\mu^2}{2} \right).
\end{align*}
If $K(\ell, \ell)$ is bounded, then the measure is finite almost surely regardless of the density $L(d \ell)$. 
The linear covariance function is unbounded. In this case, from the above equality we have
\begin{align*}
\E[\int e^{ F(\ell)} L(d \ell)] = \frac{1}{2 \pi}^{D/2} \int e^{\frac{1}{2}(1 - 1/\sigma^2)  \ell^\top \ell } d\ell < \infty,
\end{align*}
for $\sigma^2 < 1$. Putting this all together means that the Poisson rate is almost surely finite for linear
kernel when the locations are drawn from an isotropic Gaussian with variance less than $1$.  The same conditions transfer 
to the softplus variant as $\exp(x) \geq \log(1 + \exp(x))$.

\paragraph{Finiteness of the beta-Bernoulli correlated latent feature model.}
We use Proposition \ref{prop:integrability} to establish finiteness of
this measure. We note that finiteness in the number of features follows
from summability of the probability that each feature is on. 
The mean of the conditional random measure of the probabilities is
\begin{align*}
\tilde{\nu}(d\ell, da, dw, dx)  =  L(d\ell) H(da) \alpha w^{-1} (1-w)^{\alpha -1} 1\{x_i = \sigma(\sigma^{-1}(w) + F(\ell))\} dw dx.
\end{align*}
The integral of $x \wedge 1$ is the same as $x$, as $x$ is bounded by 1 due to the logistic function. Thus,
\begin{align*}
\int x&  L(d \ell) H(da) \alpha w^{-1} (1-w)^{\alpha -1} 1\{x_i = \sigma(\sigma^{-1}(w) + F( \ell))\} dw dx
\\
&= H(E) \int \alpha w^{-1} (1-w)^{\alpha -1} \sigma(\sigma^{-1}(w) + F(\ell)) dw L(d \ell).
\end{align*}
We assume that $H(E)$ is a finite, like in a probability distribution. Thus the finiteness
of this quantity only depends on the interior integral.
We can split this integral over each half of the unit interval.  The integral
of the second half is
\begin{align*}
\int &\int_{\frac{1}{2}}^1 \alpha w^{-1} (1-w)^{\alpha -1} \sigma(\sigma^{-1}(w) + F(\ell)) dw L(d\ell)
\\
& \leq 2 \int_{\frac{1}{2}}^1 \alpha (1-w)^{\alpha -1} dw = \frac{1}{2(\alpha - 1)} < \infty.
\end{align*}
This means finiteness only depends on the integral with respect to the first part of 
the unit interval. The integral over the first half is
\begin{align*}
\int &\int_0^\frac{1}{2} \alpha w^{-1} (1-w)^{\alpha -1} \sigma(\sigma^{-1}(w) + F(l)) dw L(d \ell)
\\
&= \int \int_0^\frac{1}{2} \alpha w^{-1} (1-w)^{\alpha -1} \frac{1}{1 + \frac{1-w}{w} e^{-F(l)}} dw L(d\ell)
\\
&= \int \int_0^\frac{1}{2} \alpha (1-w)^{\alpha -2} \frac{1}{\frac{w}{1 - w} + e^{-F(l)}} dw L(d \ell)
\\
&\leq \int  e^{F(\ell)} L(dl) \int_0^\frac{1}{2} \alpha (1-w)^{\alpha -2} dw = C  \int  e^{F(l)} L(d \ell),
\end{align*}
for some constant $C$. Following the same argument for CNPF above, 
this means the measure is finite when  $\int  e^{F(\ell)} L(d\ell)$ is finite.
The measure is almost surely finite for a linear
kernel when the locations are drawn from an isotropic Gaussian with variance less than $1$ and for
bounded variance covariance functions.

\paragraph{Finiteness of the gamma-Bernoulli correlated latent feature model.}
The sum of the activation probabilities can be given as
\begin{align}
Z &\triangleq\sum_{i=1}^{\infty} \frac{w_i \exp(F(\ell_i))}{w_i \exp(F(\ell_i)) + 1}
\nonumber \\
& \leq \sum_{i=1}^{\infty} w_i \exp(F(\ell_i)).
\end{align}
This is the same as the finite measure condition in CNPF.
Thus, the measure is almost surely finite for a linear
kernel when the locations are drawn from an isotropic Gaussian with variance less than $1$ and for
bounded variance covariance functions.

\subsection{Variational Inference for CNPF}
 \label{appendix:inference}

 Variational inference for CNPF maximizes the Evidence Lower BOund (ELBO).  The full ELBO for
 CNPF is
\begin{align}
  \cL =& \E_q[\log p(\alpha)] + \E_q[\log p(c)] + \E_q[\log p(s | \alpha, c)] + \sum_{k=1}^T \E_q[\log p(v_k | \alpha)] 
\nonumber  \\
       &+ \E_q[\log p(\ell_k)] + \sum_{k=1}^T \sum_{i=1}^I \E_q[\log p(a_{ki}) - \log q(a_{ki})]
\nonumber  \\
       &+ \sum_{u=1}^U  \E_q[\log p(d_u)] + \E_q[\log p(\mu_u)] + \sum_{k=1}^T \E_q[\log p(x_{ku} | d_u, v_k, s, \ell_k) - \log q(x_{ku})] 
\nonumber  \\
       &+ \sum_{i=1}^I \E_q[\log p(\tilde{y}_{uik} | x_{ku}, a_{ki}, y_{ui}) - \log q(y_{uik})],
\label{eq:full-elbo}
\end{align}
where the ELBO for
CNPF is augmented with an auxiliary variable $\tilde{y}_{ui}$ to allow for analytic
updates similar to \citet{Dunson:2005}, \citet{Zhou:2012}, and \citet{Gopalan:2014}.
We now describe its role. In variational inference, the update 
for a latent variable depends on the variational expectation of
 terms in the joint distribution where
that variable appears \citep{Bishop:2006}. The update
of the components of the base measure depends on the observation
log-likelihood
\begin{align*}
E_q[-\sum_k x_{ku} a_{ik} + y_{ui} \log(\sum_k x_{ku} a_{ik}) - \log  y_{ui}!].
\end{align*}
The second term,
\begin{align}
y_{ui} \E_q[\log(\sum_k x_{ku} a_{ik})],
\label{eq:hard-term}
\end{align}
does not have analytic form when $x_{ku}$ and $a_{ik}$ are
gamma distributed. To address this, we decompose the Poisson
observation into a sum of Poisson variables.
From the additivity of Poisson random variables, the Poisson
observation in CNPF is equivalent to
\begin{align*}
y_{ui} = \sum_{k=1}^\infty \tilde{y}_{uik},
\tilde{y}_{uik} \sim \textrm{Poisson} (x_{ku} a_{ik}),
\end{align*}
with the auxiliary $\tilde{y}_{uik}$ marginalized out. The rate of
these auxiliary Poisson is no longer a sum, so the 
variational expectations
are tractable.

In mean field variational inference,
the update to the approximating family of a latent variable 
depends on the distribution of that latent variable conditional on 
everything else \citep{Bishop:2006}. Conditional on $y_{ui}$, $a$, $z$, the vector $\tilde{y}_{ui}$ is
multinomially distributed \citep{Zhou:2012} as the following
\begin{align*}
\tilde{y}_{ui} | y_{ui}, a, x \sim \textrm{Mult} (\frac{x_{ku} a_{ik}}{\sum_{k=1}^\infty z_{ku} a_{ik}}).
\end{align*}
We introduce these auxiliary variables for only those observations that are nonzero as \myeqp{hard-term}
is zero for zero observations.

In CNPF,
there are global latent variables which are the set of global tuples, and local latent variables
which are the correlated random measure associated with each patient.
We define an equivalent
objective in terms of just the global latent variables $G$ by maximizing 
over the per-patient latent variables.

\begin{align*}
\cL^G = \textrm{max}_{\Theta} \cL(G, \Theta) = \cL(G, \Theta^*(G)),
\end{align*}
where $\Theta^*(G)$ is the setting of the per-patient parameters that maximizes
the ELBO given the global parameters $G$. To maximize $\cL^G$ we need to compute 
its gradient. By the chain rule,
\begin{align*}
\frac{\partial \cL^G}{\partial G} = \frac{\partial \cL}{\partial G}(G, \Theta^*(G)) + \frac{\partial \cL}{\partial \Theta}(G, \Theta^*(G)) \frac{\partial  \Theta^*(G)}{\partial G}(G) =  \frac{\partial \cL}{\partial G}(G, \Theta^*(G)).
\end{align*}

Thus in words, the gradient of the objective parameterized by just the global
variables is the gradient of the original objective evaluated at the maximizing
per-patient parameters given the global variables. This yields a mixed
coordinate ascent/gradient ascent maximization for this objective that allows for
parallel computation across patients. We will now 
detail all of the global gradients followed by the local coordinate updates.

\paragraph{Global gradients.}
Given the optimal variational parameters for each of the patients, we give the
gradients of the variational parameters shared across patients. The global
gradients may be prescaled by a positive definite matrix (preconditioner) for 
efficiency.

\paragraph{Natural gradient of $\alpha_{ki}^a$ and $\beta_{ki}^a$.}
The variational approximation for $a_{ki}$, the positive condition weight
in each component, is the same family as the prior, the
gamma distribution. Here $\alpha_{ki}^a$ and $\beta_{ki}^a$
represent the shape and rate of the approximation respectively. 
We compute natural gradients, which are gradients scaled by
the inverse Fisher information matrix of the variational approximation. These
gradients have been shown to have good computation properties \citep{Sato:2001, Honkela:2008, Hoffman:2013}.
The natural gradient with respect to $\alpha_{ki}^a$ and $\beta_{ki}^a$
are
\begin{align*}
\frac{\partial \cL} {\partial \alpha_{ki}^a} =& \alpha_h^a + \sum_{u=1}^U \phi_{uik} y_{ui}
\\
\frac{\partial \cL}{\partial \beta_{ki}^a} =& \beta_h^a + \sum_{u=1}^U \frac{\alpha_{ku}^x}{ \beta_{ku}^x}.
\end{align*}
From this equation computing the gradients require iterating over the entire observation matrix. For large, sparse
observation matrices, this is inefficient.
We
rewrite the gradient in terms of nonzero $y_{ui}$ as
\begin{align*}
\frac{\partial \cL} {\partial \alpha_{ki}^a} =& \alpha_h^a + \sum_{u: y_{ui} > 0} \phi_{uik} y_{ui}
\\
\frac{\partial \cL}{\partial \beta_{ki}^a} =& \beta_h^a + \sum_{u=1}^U \frac{\alpha_{ku}^x}{ \beta_{ku}^x},
\end{align*}
where we note $ \sum_{u=1}^U \frac{\alpha_{ku}^x}{ \beta_{ku}^x}$ is the same across all $i$.

\paragraph{Gradient of $\hat{s}$ and $\hat{V_k}$.}
We define the following quantity that will be useful in writing the gradient for both
 $\hat{s}$ and $\hat{V_k}$.
\begin{align*}
dw_{ku} = -\Psi(w_k) + \Psi(\alpha_{ku}^x) - \log(\beta_{ku}^x) - \hat{d_u}^\top \hat{\ell_k} - \hat{\mu_u}.
\end{align*}
Given this, the gradient of the rate of the gamma process is given by
\begin{align*}
\frac{\partial \cL}{\partial \hat{s}} = \frac{\alpha_s - 1}{\hat{s}} - c \hat{s} - \sum_{u=1}^U \sum_{k=1}^T dw_{ku} \frac{w_k}{\hat{s}},
\end{align*}
and the gradient of the sticks is
\begin{align*}
\frac{\partial \cL}{\partial \hat{V_k}} = \frac{1 - \alpha_s}{1 - \hat{V}_k} +w_{k} / \hat{V}_k  \sum_{u=1}^U dw_{ku}  - \sum_{j > k}^T dw_{ju} \frac{w_{j}}{1 - \hat{V_k}}.
\end{align*}
Positivity constraints are handled by transforming to the inverse softplus ($\log(1 + \exp(x))$) 
space where the parameters are unconstrained.
The gradient in this space follows directly from the previous equations and the chain rule. We handle all future 
positivity constraints in a similar manner. We handle the unit interval constraint on 
$V_k$ with the inverse logistic transformation.

\paragraph{Gradient of $\hat{\ell_k}$.}
The gradient of the locations
that define the correlations
is given by
\begin{align*}
\frac{\partial \cL}{\partial \hat{\ell_k}} =  -\frac{1}{\sigma_l^2}\hat{\ell}_k + \sum_{u=1}^U d_u \left( -w_k + \frac{\alpha_{ku}^x}{ \beta_{ku}^x \exp(\hat{d_u}^\top \hat{\ell_k} + \hat{\mu_u})} \right).
\end{align*}
We use the negative Hessian as the predconditioner matrix. The Hessian of the ELBO with respect to 
locations is
\begin{align}
\frac{\partial \cL^2}{\partial \hat{\ell_k} \partial \hat{\ell_k}} = -\frac{1}{\sigma_l} I_d -  \sum_{u=1}^U d_u d_u^\top \left( \frac{\alpha_{ku}^x}{ \beta_{ku}^x \exp(\hat{d_u}^\top \hat{\ell_k} + \hat{\mu_u})}\right).
\end{align}

\paragraph{Gradient of $\hat{\alpha}$.}
The gradient of the base mass of the gamma process
is given by
\begin{align*}
\frac{\partial \cL}{\partial \hat{\alpha}} = \log(\hat{c}) - (T + 1) \Psi (\hat{\alpha})  + T \Psi(\hat{\alpha} -1) +
\log(\hat{s}) + \sum_{k=1}^T \log( 1- \hat{V_k})  + (a_{\alpha}- 1) \log(\hat{\alpha}) - b_{\alpha},
\end{align*}
where $a_{\alpha}$ and $b_{\alpha}$ are respectively the 
shape and rate of the hyperprior. We set $a_{\alpha}$ to 1 and $b_{\alpha}$ to 0.01.

\paragraph{Gradient of $\hat{c}$.}
The gradient for the point estimate of the gamma process 
rate is given by
\begin{align*}
\frac{\partial \cL}{\partial \hat{c}} = \frac{\hat{\alpha}}{c} - \hat{s} + (a_c - 1) \log(\hat{c}) - b_c
\end{align*}
where $a_c$ and $b_c$ are respectively the shape and rate of the hyperprior. We set both
parameters to the same values as the prior on $\hat{\alpha}$.

\paragraph{Coordinate updates.}
To find the optimal per patient variational parameters, we iterate between
coordinate updates.

\paragraph{Coordinate update of $\tilde{y}_{ui}$.}
The variational distribution on vector of auxiliary variables $\tilde{y}_{ui}$
is the multinomial distribution. The vector $\phi_{ui}$ is the vector of probabilities to
this multinomial distribution. The optimal variational parameters given the
rest of the model is given by
\begin{align*}
\phi_{uik}^* \propto \exp(\Psi(\alpha_{ku}^x) - \log(\beta_{ku}^x) + \Psi(\alpha_{ki}^a) - \log(\beta_{ki}^a)).
\end{align*}
We again note that we only introduce auxiliary variables for $y_{ui}$ that are nonzero.  

\paragraph{Coordinate update of $x_{ku}$.}
We let the variational distribution over $x_{ku}$ be the gamma distribution.
The coordinate
updates for the shape of this variational family is
\begin{align*}
{\alpha_{ku}^x}^* = w_k + \sum_{i : x_{ui} > 0} x_{ui} \phi_{uik},
\end{align*}
and the rate is
\begin{align*}
{\beta_{ku}^x}^* = \exp(\hat{d_u}^\top \hat{\ell_k} + \hat{\mu_u}) + \sum_{i=1}^I \frac{\alpha_{ki}^a}{\beta_{ki}^a}.
\end{align*}
We note that the sum over conditions can be computed once and shared across all patients.

\paragraph{Coordinate update of $d_{u}$ and $\mu_u$.}
There is no simple closed form solution for
the coordinate update of $\hat{d_u}$. Instead, we use gradient
ascent.  
The gradient of the ELBO with respect to $\hat{d_u}$
is given by
\begin{align*}
\frac{\partial \cL}{\partial \hat{d_u}} = -\hat{d_u} + \sum_{k=1}^T  \hat{\ell_k} \left(-w_k + \frac{\alpha_{ku}^x}{ \beta_{ku}^x \exp(\hat{d_u}^\top \hat{\ell_k} + \hat{\mu_u})}\right).
\end{align*}
and the gradient for the shared Gaussian process mean is
\begin{align*}
\frac{\partial \cL}{\partial \hat{\mu_u}} = -\frac{\hat{\mu_u}}{\sigma_m^2} + \sum_{k=1}^T -w_k +  \frac{\alpha_{ku}^x}{ \beta_{ku}^x \exp(\hat{d_u}^\top \hat{\ell_k} + \hat{\mu_u})}.
\end{align*}
We terminate the procedure when the change in $\hat{d_u}$ between steps falls below a threshold or a maximum
number of iterations is reached.

\paragraph{Variational inference.}
Algorithm \ref{alg:batch} presents a variational inference algorithm using
the gradients and coordinate maximization procedures derived in the 
previous section. For the global gradients without preconditioners, we use 
RMSProp~\citep{Tieleman:2012}.
RMSProp is a per-component learning rate, which can be viewed as multiplication by a diagonal
matrix.
Formally, if $g_t$ is the gradient at iteration $t$, $\tau$ is a number in the unit interval,
and $\eta_t$ is a scalar, 
then the RMS preconditioner $\rho_t$ can be computed as
\begin{align*}
\hat{g_t^2} &= (1 - \tau) g_{t-1}^2 + \tau \textrm{diag}(g_t  g_t^\top) \\
\rho_t &=  \frac{\eta_t}{\sqrt{\hat{g_t^2}}}.
\end{align*}
Intuitively, RMSProp accounts for length scales and in the noisy setting
takes smaller steps along noisier coordinates. 

As the optimization problem for each $\Theta_u$ is independent given the global
parameters $G$, we can parallelize this step. In all of our experiments we parallelize
the maximization step across forty cores. We assess convergence using predictive perplexity
on a held out collection of patients.

\begin{algorithm}[tb]
   \caption{Variational Inference for CNPF}
   \label{alg:batch}
\begin{algorithmic}
 \STATE {\bfseries Input:} data $x$.
 \STATE {\bfseries Initialize} $G$ randomly, $t = 1$.
  \REPEAT
  \FOR {{\bf in parallel} $u=1$ {\bfseries to} $U$}
  \STATE Optimize $\Theta_u$ given $G$.
  \ENDFOR
 \STATE Follow preconditioned update for $G$.
 \UNTIL{validation perplexity stops improving. }
\end{algorithmic}
\vskip -0.05in
\end{algorithm}

\paragraph{Stochastic variational inference.}
The variational inference algorithm presented in Algorithm \ref{alg:batch}
computes the optimal local variational parameters for each patient before updating 
the variational parameters for the random variables shared across patients at each iteration. 
As the number of patients grows large, this computational cost of this becomes prohibitive.
To remedy this malady, we turn to stochastic variational inference 
\citep{Hoffman:2013}.

Stochastic variational inference works by performing stochastic optimization \citep{Robbins:1951}
on the variational objective. Stochastic optimization maximizes an objective by following a noisy
gradient which is unbiased (in expectation is the true gradient).

In stochastic variational inference, the 
noise stems from subsampling datapoints. This leads to quicker updates as the noisy
gradients are based on a fraction of the entire objective. 
Consider a patient $u$, then define the following objective
\begin{align}
\cL^u(\cG) =& \E_q[\log p(\alpha)] + \E_q[\log p(c)] + \E_q[\log p(s | \alpha, c)] + \sum_{k=1}^T \E_q[\log p(v_k | \alpha)] 
\nonumber \\
&+ \E_q[\log p(\ell_k)] + \sum_{k=1}^T \sum_{i=1}^I \E_q[\log p(a_{ki}) - \log q(a_{ki})]
\nonumber \\ 
&+ U (\E_q[\log p(d_u)] + \ E_q[\log p(\mu_u)] + \sum_{k=1}^T \E_q[\log p(x_{ku} | d_u, v_k, s, \ell_k) - \log q(x_{ku})] 
\nonumber \\
&+ \sum_{i=1}^I \E_q[\log p(y_{uik} | x_{ku}, a_{ki}) - \log q(y_{uik})]).
\label{eq:noisy_obj}
\end{align}
If we let $u \sim \textrm{Unif}(1,U)$, then $E_d[\cL^u] = \cL$. Thus,
gradient of $\cL^u$ where $u$ is uniformly drawn from $1$ to $U$ is an
unbiased gradient. The gradient of $\cL^u$ is computed by finding the
local optimal parameter for the patient $u$ and scaling it according to the
total number of patients. This objective and noisy gradient
generalizes in a straightforward manner to drawing small batches of patients.

Computationally, stochastic variational inference provides an advantage over Algorithm \ref{alg:batch},
as the slow part of Algorithm \ref{alg:batch} for large datasets is computing the optimal 
local parameters for every single datum. Algorithm \ref{alg:stoch} summarizes
stochastic variational inference for the CNPF model. 

\begin{algorithm}[tb]
   \caption{Stochastic Variational Inference for CNPF}
   \label{alg:stoch}
\begin{algorithmic}
 \STATE {\bfseries Input:} data $x$
 \STATE {\bfseries Initialize} $G$ randomly, $t = 1$.
  \REPEAT
  \STATE Draw $d \sim \textrm{Unif}(1, U).$
  \STATE Optimize $\Theta_d$ given $G$.
 \STATE Follow preconditioned update for $G$ with stochastic gradients.
 \UNTIL{validation perplexity stops improving.}
\end{algorithmic}
\vskip -0.05in
\end{algorithm}

\subsection{Stochastic Variational Inference for Baselines}
 \label{appendix:svi-baseline}
Both the HGP and the uncorrelated models are restrictions of CNPF. The HGP
is a restriction of CNPF when the locations $l$ and scalings $\mu_u$ are set
to zero, while the uncorrelated model only restricts the locations to be zero.
This means the only change required in inference is to fix the respective 
parameters to zero depending on whether inferring the HGP or the scaled HGP.

\subsection{Stochastic Variational Inference for Softplus CNPF}
 \label{appendix:svi-softplus}
The transformation distribution for Softplus CNPF is 
\begin{align*}
  x_{ui} \sim \textrm{Gamma}\left(w_i, \frac{1}{\log(1 + \exp\{F(\ell_i)\})} \right).
\end{align*}
This means the variational updates for $w$, $\ell$, $x$, $d$, and $m$ will be different. Define
$g_{uk}$ to be $\hat{d_u}^\top \hat{\ell_k} + \hat{\mu_u}$.

\paragraph{Softplus CNPF $dw$.}
The gradient of the weights is
\begin{align*}
dw_{ku} = -\Psi(w_k) + \Psi(\alpha_{ku}^x) - \log(\beta_{ku}^x) - \log(\log(1 + \exp(g_{uk}))).
\end{align*}

\paragraph{Gradient of $\hat{\ell_k}$.}
Recall $\sigma$ is the logistic function.
The gradient of the locations
is given by
\begin{align*}
\frac{\partial \cL}{\partial \hat{\ell_k}} =  -\frac{1}{\sigma_l^2} \hat{\ell}_k + \sum_{u=1}^U d_u  \frac{\sigma(g_{uk})}{\log(1 + \exp(g_{uk}))} \left(-w_k
 + \frac{\alpha_{ku}^x}{\beta_{ku}^x \log(1 + \exp(g_{uk}))} \right).
\end{align*}

\paragraph{Coordinate update of $x_{ku}$.}
The coordinate updates for the shape of $x_{ku}$ is
\begin{align*}
{\alpha_{ku}^x}^* = w_k + \sum_{i : x_{ui} > 0} x_{ui} \phi_{uik},
\end{align*}
and the rate is
\begin{align*}
{\beta_{ku}^x}^* = \log(1 + \exp(g_{uk})) + \sum_{i=1}^I \frac{\alpha_{ki}^a}{\beta_{ki}^a}.
\end{align*}

\paragraph{Coordinate update of $d_{u}$ and $\mu_u$.}
The gradient of the ELBO with respect to $\hat{d_u}$
is given by
\begin{align*}
\frac{\partial \cL}{\partial \hat{d_u}} = -\hat{d_u} + \sum_{k=1}^T  \hat{\ell_k} \frac{\sigma(g_{uk})}{\log(1 + \exp(g_{uk}))} \left(-w_k
 + \frac{\alpha_{ku}^x}{\beta_{ku}^x \log(1 + \exp(g_{uk}))} \right),
\end{align*}
and the gradient for the shared Gaussian process mean is
\begin{align*}
\frac{\partial \cL}{\partial \hat{\mu_u}} = -\frac{\hat{\mu_u}}{\sigma_m^2} + \sum_{k=1}^T \frac{\sigma(g_{uk})}{\log(1 + \exp(g_{uk}))} \left(-w_k
 + \frac{\alpha_{ku}^x}{\beta_{ku}^x \log(1 + \exp(g_{uk}))} \right).
\end{align*}
We terminate the procedure when the change in $\hat{d_u}$ between steps falls below a threshold.

\end{document}